\documentclass[oribibl]{llncs}
\usepackage{llncsdoc}

\usepackage{epsfig}
\usepackage{amssymb}
\usepackage{amstext}
\usepackage{amsmath}

\input xy
\xyoption{all}

\usepackage{algorithm}
\usepackage{algorithmic}

\usepackage{subfigure}
\DeclareGraphicsExtensions{.eps}

 \newtheorem{prop}{Proposition}
 \newtheorem{defn}{Definition}
 
 \newtheorem{exam}{Example}

  \newcommand{\G}{\mathcal{G}}
   \newcommand{\I}{\mathcal{I}}
\begin{document}

%%% ----------------------------------------------------------------------
\title{\uppercase{Knowledge Extraction and Knowledge Integration governed by {\L}ukasiewicz Logics}}
\author{Carlos Leandro}
\institute{Departamento de Matem\'{a}tica,\\
Instituto Superior de Engenharia de Lisboa, Portugal.\\
\email{miguel.melro.leandro@gmail.com}}
%
%\thanks{}
%

\maketitle
\begin{abstract}
The development of machine learning in particular and artificial intelligent in general has been strongly conditioned by the lack of an appropriate interface layer between deduction, abduction and induction \cite{Domingos09}. In this work we extend traditional algebraic specification methods \cite{Adamek94} in this direction. Here we assume that such interface for AI emerges from an adequate Neural-Symbolic integration \cite{Avila08}. This integration is made for universe of discourse described on a Topos\cite{Goldblatt06} governed by a many-valued {\L}ukasiewicz logic. Sentences are integrated in a symbolic knowledge base describing the problem domain, codified using a graphic-based language, wherein every logic connective is defined by a neuron in an artificial network. This allows the integration of first-order formulas into a network architecture as background knowledge, and simplifies symbolic rule extraction from trained networks. For the train of such neural networks we changed the Levenderg-Marquardt algorithm \cite{HaganMenhaj99}, restricting the knowledge dissemination in the network structure using soft crystallization. This procedure reduces neural network plasticity without drastically damaging the learning performance, allowing the emergence of symbolic patterns. This makes the descriptive power of produced neural networks similar to the descriptive power of {\L}ukasiewicz logic language, reducing the information lost on translation between symbolic and connectionist structures. We tested this method on the extraction of knowledge from specified structures. For it, we present the notion of fuzzy state automata,  and we use automata behaviour to infer its structure. We use this type of automata on the generation of models for relations specified as symbolic background knowledge. Using the involved automata behaviour as data sets, we used our learning methodology, to extract new insights about the models, and inject them into the specification. This allows the improvement about the problem domain knowledge.
\end{abstract}
%\begin{keyword}
%Fuzzy logics\sep {\L}ukasiewicz logic\sep Reverse Engineering\sep Symbolic Knowledge %Extraction\sep Artificial Neural Networks\sep Levenderg-Marquardt algorithm\sep %Optimal Brain Surgeon\sep Neural-Symbolic Learning Systems.
%\end{keyword}

 -----------------------------------------------------------------------
\section{\uppercase{Introduction}}

Category Theory generalized the use of graphic language to specify
structures and properties through diagrams. These categorical
techniques provide powerful tools for formal specification,
structuring, model construction, and formal verification for a wide
range of systems, presented on a grate variety of papers. The data
specification requires finite, effective and comprehensive
presentation of complete structures, this type of methodology was
explored on Category Theory for algebraic specification by
Ehresmann\cite{Ehresm68}. He developed sketches as a specification methodology of
mathematical structures and presented it as an alternative to the
string-based specification employed in mathematical logic. The
functional semantic of sketches is sound in the informal sense that
it preserves by definition the structure given in the sketch.
Sketch specification enjoy a unique combination of rigour, expressiveness and comprehensibility. They can be used for data modelling, process modelling and meta-data modelling as well thus providing a unified specification framework for system
modelling. For our goal we extend the syntax of sketch to multi-graphs and
define its models on the Topos (see e.g. for definition \cite{Johnstone02}), defined by relation evaluated in a many-valued logic. We named \emph{specification system} too our version of Ehresmanns sketch, and on its definition we developed a conservative extension to the notions of commutative diagram, limit and colimit for many-valued logic.

In this work, we use background knowledge about a problem to specify its domain structures. This type of information is assumed to be vague or uncertain, and  described using multi-diagrams. We simplify the exposition and presentation of this notions using a string-based codification, for this type of multi-diagrams, named \emph{relational specification}. We use this description for presenting structures extracted from data and on its integration.

There are essentially two representation paradigms to represent the extracted information, usually taken very differently. On one hand, symbolic-based descriptions are specified through a grammar that has fairly clear semantics. On the other hand, the usual way to see information presented using a connectionist description is its codification on a neural network (NN). Artificial NNs, in principle, combine the ability to learn and robustness or insensitivity to perturbations of input data.
NNs are usually taken as black boxes, thereby providing little insight into how the information is codified. It is natural to seek a synergy integrating the \emph{white-box} character of symbolic base representation and the learning power of artificial neuronal networks. Such neuro-symbolic models are currently a very active area of research. In the context of classic logic see \cite{Bornscheuer98} \cite{Hitzler04} \cite{Holldobler00}, for the extraction of logic programs from trained networks.  For the extraction of modal and temporal logic programs see \cite{Avila07} and \cite{Avila08}. In \cite{Komendantskaya07}  we can find processes to generate connectionist representation of multi-valued logic programs and for {\L}ukasiewicz logic programs ({\L}L) \cite{Klawonn92}.

Our approach to the generation of neuro-symbolic models uses {\L}ukasiewicz logic. This type of many-valued logic has a very useful property motivated by the ''linearity'' of logic connectives. Every logic connective can be defined by a neuron in an artificial network having, by activation function, the identity truncated to zero and one \cite{Castro98}. This allows the direct codification of formulas into network architecture, and simplifies the extraction of rules. Multilayer feed-forward NN, having this type of activation function,  can be trained efficiently using the Levenderg-Marquardt (LM) algorithm \cite{HaganMenhaj99}, and the generated network can be simplified using the "Optimal Brain Surgeon" algorithm proposed by B. Hassibi, D. G. Stork and G.J. Stork \cite{Hassibi93}.

We combine specification system and the injection of information extracted, on the specification, in the context of structures generated using a fuzzy automata. This type of automata are presented as simple process to generate uncertain structures. They are used to describe an example: where the generated data is stored in a specified structure and where we apply the extraction methodology, using different views of the data, to find new insights about the data. This symbolic knowledge is inject in the specification improving the available description about the data. In this sense we see the specification system as a knowledge base about the problem domain .

\section{\uppercase{Preliminaries}}
In this section, we present some  concepts that will be used throughout the paper.

\subsection{{\L}ukasiewicz logics}
Classical propositional logic is one of the earliest formal systems of logic. The algebraic semantics of this logic are given by Boolean algebra. Both, the logic and the algebraic semantics have been generalized in many directions \cite{Jipsen03}. The generalization of Boolean algebra can be based in the
relationship between conjunction and implication given by
$(x\wedge y)\leq z \Leftrightarrow x\leq (y \rightarrow z).
$
These equivalences, called \emph{residuation equivalences}, imply the properties of logic
operators in Boolean algebras.

In applications of fuzzy logic, the properties of Boolean conjunction are too rigid,
hence it is extended a new binary connective, $\otimes$, which is usually called \emph{fusion}, and the residuation equivalence $(x\otimes y)\leq z \Leftrightarrow x\leq (y \Rightarrow z)$ defines  \emph{implication}.

These two operators induce a structure of \emph{residuated poset} on a partially ordered set of truth values $P$\cite{Jipsen03}. This structure has been used in the definition of many types of logics. If $P$ has more than two values, the associated logics are called a \emph{many-valued logics}.

We focused our attention on many-valued logics having a subset of interval $P=[0,1]$ as set of truth values. In this type of logics the fusion operator $\otimes$ is known as a \emph{t}-norm. In \cite{Gerla00}, it is described as a binary operator defined in $[0,1]$ commutative and associative, non-decreasing in both arguments, $1\otimes x= x$ and $0\otimes x= 0$.

An example of a continuous $t$-norms is $x\otimes y=\max(0,x+y-1)$, named  \emph{{\L}ukasiewicz} $t$-norm, used on definition of {\L}ukasiewicz logic ({\L}L)\cite{Hajek95}.

Sentences in {\L}L are, as usually, built from a (countable) set of propositional variables,  a conjunction $\otimes$ (the fusion operator), an implication $\Rightarrow$, and the truth constant 0.  Further connectives are defined as:
$ \neg\varphi_1\text{ is }\varphi_1\Rightarrow 0,\;\; 1 \text{ is }0\Rightarrow 0 \text{ and }
  \varphi_1\oplus\varphi_2\text{ is }\neg\varphi_1\Rightarrow\varphi_2.$
The interpretation for a well-formed formula $\varphi$ in {\L}logic is defined inductively, as usual, assigning a truth value to each propositional variable.

The {\L}ukasiewicz fusion operator $x\otimes y=\max(0,x+y-1)$, its residue $x\otimes y=\min(1,1-x+y)$, and the lattice operators $x\vee y=\max\{x,y\}$ and $x\wedge y=\min{x,y}$, defined in $\Omega=[0,1]$ a structure of \emph{resituated lattice} \cite{Jipsen03} since:
\begin{enumerate}
  \item $(\Omega,\otimes,1)$ is a commutative monoid
  \item $(\Omega,\vee,\wedge,0,1)$ is a bounded lattice,
  and
  \item the residuation property holds, \[\text{for all }x,y,z\in
  \Omega, x\leq y \Rightarrow z \text{ iff } x\otimes y\leq z.\]
\end{enumerate}
This structure is divisible, $x\wedge y=x \otimes(x\Rightarrow y)$, and $\neg\neg x = x$. Structures with this characteristics are usually called MV-algebras \cite{Hajek98}.

However truth table $f_\varphi$ is a continuous structure, for our computational goal, it must be discretized, ensuring sufficient information to describe the original formula. A truth table $f_\varphi$ for a formula $\varphi$, in {\L}L, is a map $f_\varphi:[0,1]^m\rightarrow [0,1]$, where $m$ is the number of propositional variables used in $\varphi$. For each integer $n>0$, let $S_n$ be the set $\{0,\frac{1}{n},\ldots,\frac{n-1}{n},1\}$. Each $n>0$, defines a sub-table for $f_\varphi$ defined by $f_\varphi^{(n)}:(S_n)^m\rightarrow S_m$, given by $f_\varphi^{(n)}(\bar{v})=f_\varphi(\bar{v})$, and called the $\varphi$ \emph{(n+1)-valued truth sub-table}. Since $S_n$ is closed for the logic connectives defined in {\L}L, we define a \emph{(n+1)-valued {\L}ukasiewicz logic} ($n$-{\L}L), as the fragment of {\L}L having by truth values $\Omega=S_n$. On the following we generic call them "a {\L}L".

Fuzzy logics, like {\L}L, deals with degree of truth and its logic connectives are functional, whereas probability theory (or any probabilistic logic) deals with degrees of degrees  of uncertainty and its connectives aren't functional. If we take two sentence from $\L$ the language of {\L}L, $\varphi$ and $\psi$, for any probability  defined in $\L$ we have $P(\phi\oplus\varphi)=P(\phi)\oplus P(\varphi)$ if $\neg(\phi\otimes\varphi)$ is a boolean tautology, however for a valuation $v$ on $\L$ we have $v(\phi\oplus\varphi)=v(\phi)\oplus v(\varphi)$. The divisibility in $\Omega$, is usually taken as a fuzzy modus ponens of {\L}L, $\varphi,\varphi\rightarrow \psi\vdash \psi $, where $v(\psi)=v(\varphi)\otimes v(\psi)$. This inference is known to preserve lower formals of probability, $P(\phi)\geq x$ and $P(\varphi\rightarrow \psi)\geq y$ then $P(\psi)\geq x\otimes y$. Petr H\'{a}jek presented in \cite{Hajek952} extends this principle by embedding probabilistic logic in {\L}L, for this we associated to each boolean formula $\varphi$ a fuzzy proposition "$\varphi$ is provable". This is a new propositional variable on {\L}L, where $P(\varphi)$ is now taken to be its degree of truth.

We assume in our work what the involved entities or concepts on a UoD can be described, or characterize, through fuzzy relations and the information associated to them can be presented or approximated using sentences on {\L}L. In next section we describe this type of relations in the context of allegory theory \cite{Freyd90}.

\subsection{Relations}

A vague relation $R$ defined between a family of sets $(A_i)_{i\in Att}$, and evaluated on $\Omega$, is a map $R:\prod_{i\in Att}A_i\rightarrow \Omega$. Here we assume that $\Omega$ is the set of truth values for a {\L}L.  In this case we named $Att$ the \emph{set of attributes}, where each index $\alpha\in Att$, is called an \emph{attribute} and the set indexed by $i$, $A_i$, represents the \emph{set of possible values} for that attribute on the relation or its domain. In relation $R$ every instance $\bar{x}\in \prod_{i\in Att}A_i$, have associated a level of uncertainty, given by $R(\bar{x})$, and interpreted as the truth value of proposition $\bar{x}\in R$, in $\Omega$.

\begin{figure}
 \[
 \tiny
\xymatrix @=7pt {
&&&*++[o][F-]{G}\ar `r[rd][rd]\ar `r[rrd][rrd]\ar `r[rrrd][rrrd]&&&\\
 A_0\ar `u[urrr][urrr]&A_1\ar `u[urr][urr]& A_2\ar `u[ur][ur]&&A_3&A_4&A_5
 }
\]
\caption{A multi-arrows.}\label{graph1}
\end{figure}
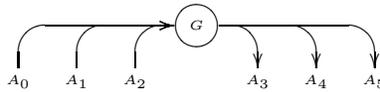

Every partition $Att_i\cup Att_a\cup Att_o=Att$, where the sets of attributes $Att_i$, $Att_a$ and $Att_o$ are disjoint, define a relation $$G:\prod_{i\in Att_i}A_i\times\prod_{i\in Att_o}A_i\rightarrow \Omega,$$ by $$G(\bar{x},\bar{z})=\bigoplus_{\bar{y}\in\prod_{i\in Att_a}A_i}R(\bar{x},\bar{y},\bar{z}),$$ and denoted by $G:\prod_{i\in Att_i}A_i\rightharpoonup\prod_{i\in Att_o}A_i$, this type of relation we call a \emph{view} for $R$. For each partition $Att_i\cup Att_a\cup Att_o=Att$ define a view $G$ for $R$, where $Att_i$ and $Att_o$ are called, respectively, the set of $G$ inputs and the set of its outputs. This sets are denoted, on the following, by $I(G)$ and $O(G)$.  Graphically a view \[_{G:A_0\times A_1\times A_2\rightharpoonup A_3\times A_4\times A_5},\] can be presented by multi-arrow on figure \ref{graph1}.

A view $S:A\rightharpoonup A$ is called a \emph{similarity relation} is
\begin{enumerate}
  \item $S(\bar{x},\bar{x})=1$ (reflexivity),
  \item $S(\bar{x},\bar{y})=S(\bar{y},\bar{x})$ (symmetry), and
  \item $S(\bar{x},\bar{y})\otimes S(\bar{y},\bar{z})\leq S(\bar{x},\bar{z})$
  (transitivity).
\end{enumerate}
We use Greek lets for similarity relation relations, and if $\alpha:A\rightharpoonup A$ is a similarity relation we write $\alpha:A$, and call to $A$ the support for similarity $\alpha$.  The similarity using $\alpha$ between to elements $\bar{x}$ and $\bar{y}$ is denoted by $[\bar{x}=\bar{y}]_\alpha$ to mean $\alpha(\bar{x},\bar{y})$.

We see a similarity relation $\alpha:A$ as a way to encoded fuzzy sets in {\L}L. We do this interpreting, for $\bar{x}\in A$, its diagonal $[\bar{x},\bar{x}]_\alpha$ as the degree of true for proposition $\bar{x}\in \alpha$. Given two elements in the support set $\bar{x},\bar{y}\in A$, we interpret  $[\bar{x},\bar{y}]_\alpha$ as the degree of true for proposition $\bar{x}=\bar{y}$ in $\alpha$. This offer us a way to evaluate the equality and de membership relation on the {\L}L.

Let $\Omega$-$Set$ be the class of views defined by relations evaluated in $\Omega$. We define a monoidal structure in $\Omega$-$Set$ for every pair of views
\[G:\prod_{i\in I(G)}A_i\rightharpoonup\prod_{i\in O(G)}A_i
\text{ and }
R:\prod_{i\in I(R)}A_i\rightharpoonup\prod_{i\in O(R)}A_i\]
we define,
\[R\otimes G:\prod_{i\in I(G)\cup I(R)\setminus O(G)}A_i\rightharpoonup\prod_{i\in O(R)\cup O(G)\setminus I(R)}A_i,\]
given, for every
\[
 (\bar{x},\bar{z})\in \prod_{i\in I(G)\cup I(R)\setminus O(G)}A_i\times\prod_{i\in O(R)\cup O(G)\setminus I(R)}A_i,
 \]
 by
 \[
(R\otimes G)(\bar{x},\bar{z})=\bigoplus_{\bar{y}\in O(R)\cap I(G)}(R(\bar{x},\bar{y})\otimes S(\bar{y},\bar{z})).
 \]
We call to this tensor product \emph{composition of view}. This operation extends composition of functions: if relation $G$ is a function between sets $A$ and $B$, and if $R$ is a function between sets $B$ and $A$, then for this two views in $\Omega$-$Set$, $G\otimes R$ is the function $R\circ G$.

\begin{figure}
\[
\tiny
 \xymatrix @=7pt {
&&&&&*++[o][F-]{G}\ar `r[rdd][rdd] &\\
&&*++[o][F-]{R}\ar `r[rrd][rrd]\ar `r[rrrd][rrrd] &&&&\\
 A_0\ar `u[urr][urr]\ar `d[drrrr][drrrr]&A_1\ar `u[ur][ur]\ar `d[drrr][drrr]& &A_2\ar `u[uurr][uurr]\ar `d[dr][dr]&A_3\ar `u[uur][uur]&A_4&A_5\\
 &&&&*+++[o][F-]{R\otimes G}\ar `r[ru][ru]\ar `r[rru][rru]&&\\
 }
\]
\caption{Composing to multi-arrows.}\label{graph2}
\end{figure}
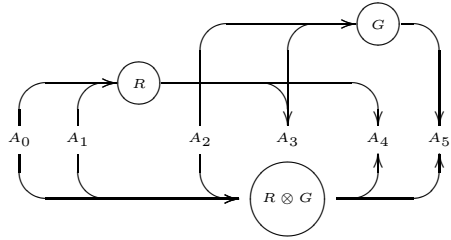

While composition between maps is a partial operator, it is defined only for componible maps, the tensor product $\otimes$ is total, it is defined for every pair of relations. In figure \ref{graph2} for two multi-arrow $R$ and $G$ representing views such that $I(R)=\{A_0,A_1\}$, $O(R)=\{A_2,A_3,A_4\}$, $I(G)=\{A_2,A_3\}$, and $O(G)=\{A_5\}$, for the resulting view $R\otimes G$ we have $I(R\otimes G)=\{A_0,A_1,A_2\}$ and $O(R\otimes G)=\{A_4,A_5\}$.

In $\Omega$-$Set$ we denote by $I:\ast\rightarrow \Omega$ the relation defined on a singleton set by $I(\ast)=1$. This relation is the identity for $\otimes$; $R\otimes I \approx I\otimes R \approx R$.

The class $\Omega$-$Set$ have a natural structure of category, having by objects $\Omega$-sets and by morphisms view, such that $R:\alpha\rightarrow \beta$ is a morphism from  $\Omega$-set $\alpha:A$ to $\beta:B$ if $R:A\rightharpoonup B$ and
\begin{enumerate}
  \item $\alpha \otimes R \leq R$, and
  \item $R \otimes \beta \leq R$.
\end{enumerate}
Note that every object $\alpha:A$ have by identity the relation $\bar{\alpha}:A\rightharpoonup A$ defined by reflexive the close of $\alpha$, define making $\bar{\alpha}(\bar{x},\bar{x})=1$.

 The category $\Omega$-$Set$ is a symmetric monoidal closed category  \cite{Borceux94}, where the tensor product of $\Omega$-sets is given for $\alpha:A$ and $\beta:B$ by
 $$\alpha\otimes\beta:A\times B$$
 defined  $$(\alpha\otimes\beta)(a,b)=\alpha(a)\otimes\beta(b).$$
 This can be used to describe a functor
 \[
 \alpha\otimes\_ : \Omega-Set \rightarrow \Omega-Set,
 \]
 given for morphisms  $R:\gamma \rightharpoonup \beta$, with support $f:A \rightharpoonup B$, by
 \[
      \alpha\otimes f:\alpha\otimes\gamma\rightharpoonup \alpha\otimes\beta
 \]
 having by support $\alpha\otimes f:A\times C\rightharpoonup A\times B$, described by
      \[
      (\alpha\otimes f)(a,c,a',b)=\alpha(a,a')\otimes f(c,b).
      \]

Functor $\alpha\otimes\_$ have by left adjunct a functor
      \[
      \alpha\multimap\_ : \Omega\text{-}Set \rightarrow \Omega\text{-}Set,
      \]
defined for $\Omega$-sets $\beta:B$ by
      \[
      \alpha\multimap \beta:[A,B],
      \]
construct as the internalization  for $\Omega$-set $Hom$ \cite{Clementino04}
      \[
      (\alpha\multimap \beta)(t,h)=\bigvee_{b_0,b_1}\bigoplus_{a}( \alpha(a,a)\otimes t(a,b_0)\otimes h(a,b_1)\otimes \beta(b_0,b_1)),
      \]
for relations $f:\gamma\rightharpoonup \beta$, with support $f:C\rightharpoonup D$, we have
      \[
      (\alpha\multimap f):(\alpha\multimap \gamma)\rightharpoonup (\alpha\multimap \beta),
      \]
a relation with support $\alpha\multimap f:[A,C]\rightharpoonup [A,B]$, described by
       \[
       (\alpha\multimap f)(h,g)(a,c,a',b)=h(a,c)\otimes g(a',b)\otimes \alpha(a,a').
       \]
This adjunction $\alpha\otimes\_ \;\vdash\; \alpha\multimap\_ $ have by unit \cite{Borceux94} the natural transformation, $\lambda$ defined for each $\Omega$-set $\gamma:C$, by a multi-morphism
       \[
       \lambda_\gamma:(\alpha\multimap \gamma)\otimes \alpha \rightharpoonup \gamma,
       \]
with support $\lambda_\gamma:[A,C]\times A \rightharpoonup C$, by
       \[
       \lambda_\gamma(h,a,b)=h(a,b),
       \]
the relation $h$ evaluation evaluated in $(a,b)\in A\times B$.

The $\alpha\multimap \beta:[A,B]$ reflexive closure  defines a similarity relation in $[A,B]$, we use this relation in the following to quantify the similarity between relation form $S$ and $B$, and we call them \emph{power similarity relation}. In the follow we use this relation to compare models or on the quantification of model quality.

Two views $R$ and $G$, in $\Omega$-$Set$, are called \emph{independents} if $R\otimes G=G\otimes R$. By this we mean what the $R$ output not depend on $G$ inputs and the $G$ output not depend on $R$ input. Given a view $R:A\rightharpoonup B$,  we define projections $R_A:B\rightarrow\Omega$ and $R_B:B\rightarrow\Omega$, respectively, by $R_A(\bar{b})=\bigoplus_{\bar{a}\in A}R(\bar{a},\bar{b})$ and $R_B(\bar{a})=\bigoplus_{\bar{b}\in B}R(\bar{a},\bar{b})$. In the following we used $R(\bar{a},\_)$ to denote the relation defined from $R$ by fixing a input vector $\bar{a}\in A$, $R(\bar{a},\_)(\bar{b})=R(\bar{a},\bar{b})$.

\subsection{Inference}
Generically inference is a process used to generate now facts based on known facts. On the context of multi-valued logic, the inference allows fining the degree of two for a new proposition based on the known degree of truth for propositions \cite{hajek97}. This inference can be described using the composition operator defined in $\Omega$-$Set$ \cite{Zadeh75}.The \emph{syllogism} describe by the rule:
\begin{center}
             \begin{tabular}{l}
                $R$: If $a\in \alpha$ then $b\in \beta$\\
                $S$: If $b\in \beta$ then $c\in \gamma$ \\
                \hline
                $R\otimes S$: If $a\in \alpha$ then $c\in \gamma$ \\
              \end{tabular}
\end{center}
This rule is interpreted saying that:
If
\begin{enumerate}
  \item $R(a,b)\geq([a]_\alpha\Rightarrow [b]_\beta)$, and
  \item $S(b,c)\geq([b]_\beta\Rightarrow [c]_\gamma)$
\end{enumerate}
then
\begin{center}
  $(R\otimes S)(a,c) \geq [a]_\alpha\Rightarrow [c]_\gamma$.
\end{center}
This gives us a lower bond for degree of truth. However this strategy works better on the version of \emph{Modus Ponens}:
\begin{center}
             \begin{tabular}{l}
                $R$: $a\in \alpha$ \\
                $S$: If $a\in \alpha$  then $b\in \beta$ \\
                \hline
                $R\otimes S$: $a\in \alpha\; \wedge\; b\in \beta$ \\
              \end{tabular}
\end{center}
Since {\L}L is a divisible logic we can write:
\begin{center}
\begin{tabular}{rcl}
  $(R\otimes S)(a,b)$ & $=$ & $R(a,a)\otimes S(a,b)$ \\
                      & $=$ & $[a]_\alpha \otimes ([a]_\alpha\Rightarrow [b]_\beta)$  \\
                      & $=$ & $[a]_\alpha\wedge [b]_\beta$\\
\end{tabular}
\end{center}
Applying this rule to a simple relation  $H:\alpha\rightarrow \beta$, we have
\begin{center}
             \begin{tabular}{l}
                $R$: $a\in \alpha$ \\
                $S$: If $a\in \alpha$  then $(a,b)\in H$ \\
                \hline
                $R\otimes S$: $a\in \alpha\; \wedge\; (a,b)\in H$ \\
              \end{tabular}
\end{center}
since $[a]_\alpha=\bigoplus_b H(a,b)$, the degree of truth of $a\in \alpha\; \wedge\; (a,b)\in H$ is the degree of truth for $(a,b)\in H$, then:
\begin{equation}
 [a]_\alpha \otimes ([a]_\alpha\Rightarrow H(a,b))= H(a,b).
\end{equation}
We simplified this excretion defining
\begin{equation}
H(\beta | a)(b)=[a]_\alpha\Rightarrow H(a,b),
\end{equation}
and we write
\begin{equation}
 [a]_\alpha \otimes H(\beta | a)(b)= H(a,b).
\end{equation}
In this context $H(\beta | a)(b)$ is interpreted as the degree of truth for the proposition:
\begin{center}
''A class associated by relation $H$ is $b$, if its input is $a$'',
\end{center}
 given by the result for the evaluation of $(a,b)$ by $H$, conditionated to the degree belonging of $a$ on $f$ input domain. In classic logic, when $A$ is finite, this is express by $\forall a\in \alpha: H(a,b)$.

\begin{prop}[Bayes Rule on {\L}L]
Given a faithful view ${R:A\rightharpoonup B,}$ and $\bar{a}\in A$ and $\bar{b}\in B$ from $\Omega$-Set. The equations
 $$R(\bar{a})_B\otimes x = R(\bar{a},\_)\text{ and }R(\bar{b})_A\otimes x = R(\_,\bar{b}),$$
have by solution, relation $R(\_|\bar{a})$ and $R(\_|\bar{b})$, respectively, defined by $R(\_|\bar{a})=R(\bar{a})_B\Rightarrow R(\bar{a},\_)$ and $R(\_|\bar{b})=R(\bar{b})_A\Rightarrow R(\_,\bar{b})$.
\end{prop}
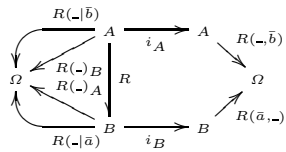
\begin{figure}[h]
\[
\tiny
\xymatrix @=9pt {
       &&A \ar[lld]^{R(\_)_B}\ar `l[lld]_{R(\_\mid \bar{b})}[lld]\ar@_{->}[dd]^R\ar[rr]_{i_A}&&A\ar[rd]^{R(\_,\bar{b})}&&\\
\Omega &  & &&&\Omega\\
       &&B \ar[llu]_{R(\_)_A}\ar[rr]_{i_B}\ar `l[llu]^{R(\_\mid \bar{a})}[llu]&&B\ar[ru]_{R(\bar{a},\_)}&&
 }
\]
\caption{Functional dependencies, where arrows $i_A$ and $i_B$ denote the identity relation.}
\end{figure}

We use this rule to solve inference problems in $\Omega$-Set. Given two compatible views $R:A\rightharpoonup B$ and $G:B\rightharpoonup C$, i.e. such that the output attributes for view R are the input attributes for G. For observable descriptions $\bar{a}\in A$ and $\bar{c}\in C$, we have
\begin{center}
\begin{tabular}{rcl}
  $R(\bar{a})_B\otimes (R\otimes G)(\_|\bar{a})(\bar{c})$ & = & $(R\otimes G)(\bar{a},\bar{c})$ \\
   & = & $\bigoplus_{\bar{b}} R(\bar{a},\bar{b})\otimes S(\bar{b},\bar{c})$\\
   & = & $\bigoplus_{\bar{b}} R(\bar{a})_B\otimes R(\_|\bar{a})(\bar{b})\otimes G(\bar{b},\bar{c})$,\\
\end{tabular}
\end{center}
then
\[
(R\otimes G)(\_|\bar{a})(\bar{c})=R(\bar{a})_B\Rightarrow (R(\bar{a})_B\otimes\bigoplus_{\bar{b}}  R(\_|a)(b)\otimes S(b,c)),
\]
i.e.
\[
(R\otimes G)(\_|\bar{a})=\bigoplus_{\bar{b}} R(\_|\bar{a})(\bar{b})\otimes S(\bar{b},\_).
\]
When views $R:A\rightharpoonup C$ and $G:B\rightharpoonup D$ are independent we have
\[(R\otimes G)(\_|\bar{a},\bar{b})(\bar{c},\bar{d})= R(\_|\bar{a})(\vec{c}) \otimes S(\_|\bar{b})(\bar{d}).\]
Naturally, if $C=D$ we write $(R\otimes G)(\_|\bar{a},\bar{b})(\bar{d})$ for $(R\otimes G)(\_|\bar{a},\bar{b})(\bar{d},\bar{d})$.

\subsection{Limit sentences and colimit  sentences}
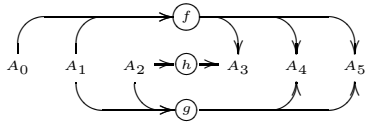
\begin{figure}[h]
\[
\tiny
\xymatrix @=7pt {
&&&*+[o][F-]{f}\ar `r[rd][rd]\ar `r[rrd][rrd]\ar `r[rrrd][rrrd]&&&\\
 A_0\ar `u[urrr][urrr]&A_1\ar `u[urr][urr]\ar `d[drr][drr]& A_2\ar `d[dr][dr]\ar[r]&*+[o][F-]{h}\ar[r] &A_3&A_4&A_5\\
 &&&*+[o][F-]{g}\ar `r[rru][rru]\ar `r[rrru][rrru]&&&
 }
\]
\caption{Example of multi-graph defined by three multi-arrows}\label{multidiagram}
\end{figure}

A multi-arrow defines a link between a set of input nodes and a set of output nodes, we can see an example of this on figure \ref{graph1}. We can use multi-arrows to generalize the notion of arrow in a graph. This allows the definition of a multi-graph as a set of nodes linked together using multi-arrows. Examples of multi-graphs can be seen on figures \ref{graph2} and \ref{multidiagram}. A multi-diagram in $\Omega$-$Set$, defined having by support a multi-graph $\G$, is a multi-graph homomorphism $D:\G\rightarrow \Omega\text{-}Set$, where each node in $\G$ is mapped to a $\Omega$-set $\alpha:A$, and each multi-arrows in $\G$ is mapped to a relation view. In this sense, every set of views in $\Omega$-$Set$ defines a multi-diagram, having by support the multi-graph where the selected views are multi-arrows, and the $\Omega$-set used on this views as nodes.

The classically definition of limit for a diagram, in the category of sets, can been as a way to internalize the structure of a diagram in form of a table \cite{Borceux94}. Given a diagram $D:\G\rightarrow Set$ with vertices $V=\{a_i\}_{i\in I}$ and arrows $A=\{f_j\}_{j\in J}$, its limit is a table or a subset of the cartesian product $\prod_{i\in I}D(a_i)$ given by
\begin{equation}
Lim\;D=\{(\ldots,x_i,\ldots,x_j,\ldots)\in \prod_{i}D(a_i):\forall_{f:a_i\rightarrow a_j}D(f)(x_i)=x_j\}.
\end{equation}
were the relation is evaluated on classic logic.

We present as limit for a multi-diagram $D:\G\rightarrow \Omega\text{-}Set$ a conservative extension from the classical limit definition. Let $D:\G\rightarrow \Omega\text{-}Set$ be a multi-diagram with vertices $(v_i)_{i\in L}$. The \emph{limit} of diagram $D$ is a relation denoted by $Lim\;D$, and defined as \[Lim\; D: \prod_{i\in L} D(v_i) \rightarrow \Omega,\]
such that
\[
(Lim\;D)(\ldots,\overline{x}_{i},\ldots,\overline{x}_{j},\ldots)=\bigotimes_{f:v_i\rightharpoonup v_j\in \G} D(f)(\overline{x}_{i},\overline{x}_{j}).
\]
The limit for multi-diagram on figure \ref{multidiagram} is the relation $
Lim\;D:A_0\times \ldots \times A_5\rightarrow \Omega
$
given for every $(a_0,\ldots,a_5)\in A_0\times \ldots\times A_5$ by
$$(Lim\;D)(a_0,a_1,a_3,a_4,a_5)= f(a_0,a_1,a_3,a_4,a_5)\otimes g(a_1,a_2,a_4,a_5)\otimes h(a_2,a_3).$$

In this sense for parallel views $R,S:X\rightharpoonup Y$, they define a multi-diagram, and its limit is the relation
$$Lim(R=S):X\times Y\rightarrow \Omega,$$
given by\\
$$
Lim(R=S)(x,y)=R(x,y)\otimes S(x,y).
$$
This relation is denoted by $[R=S]$ and usually called, on Classic logic, $R$ and $S$ \emph{equalizer}. If $R:X\rightharpoonup U$ and $S:Y\rightharpoonup U$ are views its \emph{pullback}, denoted by $R\otimes_US$ is defined by the limit
$$Lim(R\otimes_U S):X\times U\times Y\rightarrow \Omega,$$
given by
$$
Lim(R\otimes_U S)(x,u,y)= R(x,u)\otimes S(y,u).
$$
Given a family of views, having the some output, $(R_i:X_i\rightharpoonup U)_{i\in L},$ its \emph{wide-pullback} is the relation $Lim(\otimes_U R_i):\otimes_{i\in L}R_i$.

\begin{defn}[$\lambda$-Limit]\label{def:lambdaLim}
A relation $R$ described in {\L}L, is the $\lambda$-limit for a multi-diagram $D$  if $R:A\rightarrow \Omega$ is $\lambda$-similar to $Lim\;D:A\rightarrow \Omega,$ i.e if
\begin{equation}
(R\multimap Lim\;D)\geq \lambda,
\end{equation}
when this is the case we write
\begin{equation}
R=Lim_\lambda\;D.
\end{equation}
\end{defn}

We used the definition of limit to extend the notion of commutative diagram. The idea was to characterize a commutative diagrams using its internalization on a table.

\begin{defn}[Commutativity ]\label{Comutatividade Pdiagramas}
If $D:\G\rightarrow \Omega\text{-}Set$ is a multi-diagram with vertices in $V$, and associated  $\Omega$-sets $(\alpha_i)_{i\in V}$, where
we selected a set $s(D)$ of input vertices. Assuming that the sub-graph of $\G$ defined by vertices $s(D)$ is acyclic and
that $P$ is the Cartesian product defined by each vertices on $D$ with not belong to $s(D)$.

The multi-diagram $D$ is commutative with inputs if $s(D)$ if  \begin{equation}
\bigvee_{\bar{n} \in P}(Lim\;D)(\bar{s},\bar{n})=\bigvee_{\bar{n}\in V}(\bigotimes_i\;\alpha_i)(\bar{s},\bar{n}),
\end{equation}
for every $\bar{s}\in \prod_{i\in s(D)}D(i)$. A diagram is $\lambda$-commutative if
\begin{equation}
\left(\bigvee_{\bar{n}\in V}(Lim\;D)(\_,\bar{n})\multimap \bigvee_{\bar{n}\in V}(\prod_i\;D(i))(\_,\bar{n})\right)\geq\lambda,
\end{equation}
\end{defn}

Limits, colimits and commutativity  can be used on the specification of structures \cite{Adamek94}. We use the conservative extensions to this notions for the detrition of fuzzy structures. However the notion of
colimit is more difficult to present generically. The construction of a colimit reduces to that of
two coproducts and a coequalizer, siting \cite{Borceux94}, in the category of sets governed by classic logic the explicit description of a coequalizer is generically very technical since it involves the description of the equivalence relation generated by a family of pairs. This complexity is incased when we extend this notion to relations evaluated on multi-valued logics. We present bellow two examples.

The coproduct  of $\Omega$-sets $\alpha:A$ and $\beta:B$ is a relation $R$ having by support set $A\coprod B$ given by
\begin{equation}
R(a,a')=[a=a']_\alpha\oplus [a=a']_\beta.
\end{equation}
Where, for simplicity, we assume  what relations  $\alpha$ and $\beta$ assume the value $0$ when are evaluating pairs outside its support sets. We denote the coproduct for $\alpha:A$ and $\beta:B$ by $\alpha\oplus\beta:A\coprod B$.

The diagram defined by a parallel pair of multi-morphisms $f:\alpha:A\rightarrow \beta:B$ and $g:\alpha:A\rightarrow \beta:B$ have by  colimite a $\Omega$-set, with support  $A\coprod B$, given by
\begin{center}
$
R(a,a')= \begin{array}{cl}
                 & \bigoplus_{b,b'\in B}f(a,b)\oplus f(a',b')\oplus [b=b']_B\\
               \oplus  & \bigoplus_{b,b'\in B}g(a,b)\oplus g(a',b')\oplus [b=b']_B\\
               \oplus & \bigoplus_{b,b'\in A}f(b,a)\oplus g(b',a')\oplus [b=b']_A \\
               \oplus & [a=a']_A \\
               \oplus & [a=a']_B \\
              \end{array},
$
\end{center}
where $a,a'\in A\coprod B$, for simplicity, we assume  what relations  $\alpha$ and $\beta$ assume the value $0$ when are evaluated on pairs outside its support sets.

\subsection{Concepts}
We describe a table or a concept using relation views. A table or a concept description using values in the family $(A_\alpha)_{\alpha\in Att}$, for attributes $Att$,  is a view
$
{R:O\rightharpoonup\coprod_{\alpha\in Att}A_\alpha},
$
where $O$ is a set of keys identifying concept instances. We use $R(o,\alpha=x)=\lambda$ to denote that, in instance $o\in O$, the uncertainty of an attribute $\alpha$ to be equal to value $x\in A_\alpha$ is $\lambda$. This mean that, in an instance, an attribute may assume different values, associated with different uncertain levels expressed by truth values. When we have $R(o,\alpha=x)\geq\lambda$, for every entity $o\in O$, we write $R(\alpha=x)\geq\lambda$ or just $\alpha\sim_\lambda x$ in $R$.

A concept description have different presentations, corresponding to each of the perspectives taken to data. Each partition $Att=V\cup U$, defines a perspective through the view $
_{R_{V,U}:O\times\prod_{\alpha\in V}A_\alpha\rightharpoonup\coprod_{\alpha\in U}A_\alpha},
$ given by $$_{R_{V,U}(o,\overline{\alpha=x},y)=\bigotimes_{\alpha\in V,x\in A_{\alpha}}R(o,\alpha=x)\otimes R(o,y)},$$ where $\overline{\alpha=x}$ abbreviates the tuple defined using family $(\alpha=x)_{\alpha\in V,x\in A_{\alpha}}$.

Relation between information on a data set can be defined as a diagram $D$, from a multi-graph $\G$ to $Set(\Omega)$, where each multi-arrow is mapped to a view of a concept description. Every multi-graph homomorphism $I:\I\rightarrow \G$ defines a query in the structure $D$, having by answer the concept description defined by $Lim\;D\circ I$. Where $D\circ I$ denotes the composition between graph homomorphisms.

If we assume that $\Omega=[0,1]$, given a pair of concept presentations defined using a finite set of keys $O$, \[_{R_0,R_1:O\times\prod_{\alpha\in V}A_\alpha\rightharpoonup\coprod_{\alpha\in U}A_\alpha},\]
we measure the similarity between this two views using relation
\[\Gamma(R_0,R_1)=e^{-\frac{1}{|O|}\sum_{\bar{x}\in\prod_{\alpha\in Att}
A_\alpha}\neg(R_0(\bar{x})\Leftrightarrow R_1(\bar{x}))},\] where $|O|$ is the number of keys. Relation $\Gamma$ is a \emph{similarity relation} between pairs of concept described using a tuple in $\prod_{\alpha\in Att}
A_\alpha$ since:
\begin{enumerate}
  \item $\Gamma(R_0,R_0)=1$ (reflexivity),
  \item $\Gamma(R_0,R_1)=\Gamma(R_1,R_0)$ (symmetry), and
  \item $\Gamma(R_0,R_1)\otimes\Gamma(R_1,R_2)\leq\Gamma(R_0,R_2)$
  (transitivity).
\end{enumerate}
The transitivity is a consequence of, in any ML-algebra $\Omega$, for all  $\lambda_0,\lambda_1,\lambda_2\in
\Omega$,$(\lambda_0\Rightarrow \lambda_1)\otimes(\lambda_1\Rightarrow \lambda_2)\leq \lambda_0\Rightarrow \lambda_2$.
When $\Gamma(R_0,R_1)=\lambda$ we write $R_0\sim_\lambda R_1$ and we say that, $R_0$ is $\lambda$-similar to $R_1$, for $R_0\sim_1 R_1$ we write $R_0=R_1$.

We named this similarity measure of \emph{exponential similarity}. In the literature we can find other measures for relation similarity measurement. Like the \emph{inf-similarity}, used in Possibilistic logic,
\[_{\Gamma(R_0,R_1)=\bigwedge_{\bar{x}\in\prod_{\alpha\in Att}
A_\alpha}(R_0(\bar{x})\Leftrightarrow R_1(\bar{x})),}\]
or the \emph{and-similarity}, used in Boolean logic,
\[_{\Gamma(R_0,R_1)=\bigotimes_{\bar{x}\in\prod_{\alpha\in Att}
A_\alpha}(R_0(\bar{x})\Leftrightarrow R_1(\bar{x})),}\]
however these relations are to crispy for model evaluation. We need to be able to quantify  who models are similar to a concept description.

This notion is fundamental to make fuzzy some key concepts of relational algebra, useful on data structure specification. Example of this is the description of a "is\_a" relation evaluated on multi-valued logic. For that, let $R:A\rightharpoonup B$ be a concept description, here we assume the existence of a similarity $\Gamma_A$ defined in $A$. The concept description $R$ defines a "is\_a" relation, for similarity $\Gamma_A$, if sentence
\[
_{R(a_0,b)\otimes R(a_1,b) \Rightarrow \Gamma_A(a_0,a_1),}
\]
have by truth-value 1, for every $a_0,a_1\in A$ and every $b\in B$, i.e. $$_{\bigotimes_{a_0,a_1\in A}\bigotimes_{b\in B}(R(a_0,b)\otimes R(a_1,b) \Rightarrow \Gamma_A(a_0,a_1))=1.}$$
In this sense we call to view $R$ a \emph{mono-view} or a \emph{clustering}.

Views $R:A\rightharpoonup B$ such that $_{\bigotimes_{b\in B}\bigoplus_{a\in A}R(a,b)}=1$, are called \emph{epi-views}.

\section{\uppercase{Specification system}}
All the widely used data specification mechanisms (like Entity Relationship Model \cite{Chen76}, the Fundamental Data Model \cite{Ship81}, the Generic Semantic Model \cite{Abiteboul95}), OOA\&D-schemas in a million of versions and UML which itself comprises a host of various
notations, have a strong graphical component. They are essentially graphs with special markers in them. Usually the semantics of these markers is defined in an ad-hoc and sometimes non-formal way. An important component of the mathematical structure that will be used to formalized knowledge, are multi-graph homomorphisms into the class $Set(\Omega)$ of relations views. When specifying an information system, it will be necessary to formulate constrains on such graph homomorphisms. A identical notion of a specification whose models are graphs homomorphisms into a category theory are known on the category community under the name of sketch. Sketches where invented by Charles Ehresmann and can be perceived as a graphic based logic, which formalizes in a precise and uniform way the semantic of graph with marks \cite{piessen00}. For our propose we extended Ehresmann's sketch to formalize a graphic based fuzzy logic. We named, this mathematical structure, \emph{specification system}. On information specification, specification system will be used to specify finite fuzzy structures. Semantic data specification have been used for may years in the early stages of database design, and they have become key ingredients of object-oriented software. The goal of a semantic data specification is to build a mathematical abstraction of a small part of the real word. This small part of the word is usually called the \emph{universe of discourse} (UoD) in the database literature. The models of the data specification are possible states of the UoD, and will be the structures stored in an information system. The mathematical structure that will be used to describe the UoD are finite models of specification systems together with a labeling of all the elements of these models.

Hence a data specification will consist of two parts. The first part will be a specification system, and it describes the fuzzy structure and the interdependencies of the various entities about which we want to store information. The second part indicates what kind of information we want to store about each type of entity: for each type of entity, we give its set of possible attribute values, i.e. the set of all possible labels that an entity of the given type can have. The structure defined, by specification system $S$ and model $M$, we named \emph{the semiotic} $(S,M)$.

By a \emph{specification system} $S$ we understood as a structure $S=(\G,C,L,coL)$, where $\G$ is a multi-graph, $C$ is a set of pairs $(G,\lambda)$, and  $L$ and $coL$ are sets of tuples $(f,G,i(G),o(G),\lambda)$, such that $f$ is a multi-arrow in $\G$, $D\subset\G$ is a multi-graph, $\lambda\in \Omega$ and $i(G)$ and $o(G)$ are sets of nodes from $\G$.

Given a specification system $S=(\G,C,L,coL)$ a model for $S$ is a diagram $M:\G\rightarrow Set(\Omega)$, mapping multi-arrows to concept description, such that:
\begin{enumerate}
  \item for every $(G,\lambda)\in C$, $M(G)$ is  $\lambda$-commutative,
  \item for every $(f,G,i(G),o(G),\lambda)\in L$, $M(f)$ have by input $M(i(G))$ and by output $M(o(G))$ and is $\lambda$-equivalent to relation $Lim\;M(G)$, and
  \item for every $(f,G,i(G),o(G),\lambda)\in L$, $M(f)$ have by input $M(i(G))$ and by output $M(o(G))$ and is $\lambda$-equivalent to relation $coLim\;M(G)$.
\end{enumerate}
The pair $(S,M)$ defined by a model $M:\G\rightarrow Set(\Omega)$ for a specification system $S$ is called a \emph{semiotic}, where the multi-graph $\G$ describes a \emph{library of components}.

A specification system $S$, if consistent, describes the fuzzy structure for a class of UoD. If the set of all models for specification system $S$ is denoted by $Mod(S)$, every $M\in Mod(S)$ can be seen as a system state. Two states $M_0,M_1\in Mod(S)$ have similar structures and the specification $S$ can be enriched with now knowledge to increase the dissimilarity between the states. The knowledge need to distinguish between $M_0$ and $M_1$ can be extracted querying the state $M_0$, trying to find its particularities.

For that, every multi-graph homomorphism  $I:\I\rightarrow\G$ defines a query to a model $M:\G\rightarrow Set(\Omega)\in Mod(S)$. And this query $I$ have by answer the relation given by $$Lim\;M\circ I,$$ and each of its views can be used as a data set, usable to feed a data mining processes, to extracted insights about the model.

\subsection{Knowledge integration via specification systems}
Specification systems are graphic specifications formalisms (its components are multi-graphs and markers in these graphs) and can be taken as repositories of knowledge about the $UoD$. They are described using a rigorous graphic language with a precise semantic, where we have a methodology to querying its models. Your goal is the enrichment of this structure with knowledge extracted from one of its models. We simplify this process by expressing constraints of first-order logic formulae into a graphical marker.

To be able to do that, we consider the multi-graph used on the definition of the specification system component library as a presentation of a first-order many-sorted signature: nodes of the graph are interpreted as data sorts, and multi-arrows are interpreted as relations evaluated in $\Omega$. A structure for this signature is exactly a multi-graph homomorphism from $\G$ to $Set(\Omega)$. Every formula for this signature are particularly simple, since there are only relations. A formula $R(x_1,x_2,\ldots,x_n)$ is defined through a multi-graph homomorphism  $R:\I\rightarrow\G$ and its interpretation in model $M$, is the relation $Lim\;M\circ R$. Where the interpretation of each initial and terminal node of the multi-diagram $R$ defines the sorts for variables $x_1,x_2,\ldots,x_n$.

Every atomic formula is a relation $R(x_1,x_2,\ldots,x_n)$. The finite conjunction of atomic formulas on variables $x_1,x_2,\ldots,x_n$, using relations $R_1,R_2,\ldots,R_m$ is denoted by: $(R_1\otimes R_2\otimes\ldots\otimes R_m)(x_1,x_2,\ldots,x_n)$ and it is interpreted as the relation $Lim\;M\circ(R_1\otimes R_2\otimes\ldots\otimes R_m)$. In this sense, every disjunctive formula defined using relation can be interpreted as the limit of a finite diagram.

The translation in the other direction is also simple. Given a limit mark $(f,G,i(G),o(G),\lambda)$ in $S$, the meaning for sign $f$, is the interpretation of a disjunctive formula defined trough the interpretation of each multi-arrows used on the definition of multi-graph $G$ having its interdependencies (gluing order in $G$) defined using variable repetition.

Similarly, every conjunctive formula defined using relation on semiotic $(S,M)$ can be interpreted as the colimit of a finite diagram. For colimit makes \\ $(f,G,i(G),o(G),\lambda)$ is $S$, the interpretation for $f$, can be defined as the interpretation for a conjunctive formula defined using each multi-arrow in $G$ .

Given a semiotic $(S,M)$ every formula $\varphi$, extracted from model $M$ using a query $I$, can be described by a set of limit marks and colimit marks. This allows the enrichment of specification system $S$, defining a new system $S'$, such that $Mod(S')\subseteq Mod(S)$ and $M\in Mod(S')$.

However, it is known from \cite{Adamek94} what where are structures specifiable using sketches but not in first-order logic. Then first-order logic have less expressive power as specification systems. This result allows to use a mixture of limits, colimits and formulas in the UoD specification and be sure that the resulting specification can always be translated to a specification system \cite{piessen00}.

\subsection{A specification system description}

A specification system is by nature a graphic specification described, however, some times like in this exposition, it is preferable a description for the specification system  using a string-based presentation. For this we used a string-based codification for specification systems  named  \emph{relational specification}, generalizing the notion of essentially algebraic specification: the original idea goes back to Freyd \cite{Freyd72} and having the same expressive power as finite limit sketches \cite{Barr90}. The essentially algebraic fragment, are interesting computer sciences mainly because theories, of many kinds of specification formalisms, are in fact initial algebras for some essentially algebraic specification. A number of proof systems for essentially algebraic specification have been introduced \cite{Makkai93}.

Recall that an ordering on a set $X$ is well-founded iff every strictly decreasing chain of elements of $X$ must be finite. A relational specification consists of:
\begin{enumerate}
  \item A set of sorts.
  \item A set of relational signs, with a well-founded ordering (called a \emph{library of components}).
  \item A set of diagrams defined using relational views, with a well-founded ordering (called a \emph{a set of diagrams}).
  \item A set of condition build on relational signs and diagrams.
\end{enumerate}
Every relation sign or view sign $\omega$ have an arity and a set $Def(\omega)$ of relations with the same arity as $w$, called the \emph{set of domain condition}.

For our propose sorts are nominal and finite, and we list its possible values by writing   $A:\{a_1,a_2,\ldots,a_t\},$. These are the basic structures for UoD specification, used on the description of relations and views.
Every relation have an arity described by a list of data sorts. We denote a relation symbol $R$, with arity the list of sorts $A_1,A_2,\ldots,A_n$, as $R:A_1, A_2, \ldots, A_n$. Interpreted as a concept description  $M(R):O\rightharpoonup\coprod_i A_i$.

A view of $R$ having by source arity the list $A_1,A_2,\ldots,A_n$ and by target arity $B_1,B_2,\ldots,B_m$ is denoted by:
 \begin{center}
\tiny
\begin{tabular}{rll}
  $R:\{$ \\
           &$A_1, A_2, \ldots, A_n\rightharpoonup B_1, B_2, \ldots, B_m$;         & \\
          \} \\
\end{tabular}
\end{center}
These are interpreted as concept descriptions  $M(R):O\times\prod_i A_i\rightharpoonup \coprod_j B_j$.

The relationship between relations and views may be defined using marked diagrams. A diagram is described by a set of views where we marked some of its sorts as input or output sort. We write
\begin{center}
\tiny
\begin{tabular}{rll}
  $D:\{$ &$A_1,A_2,\ldots,A_n\rightharpoonup  B_1,B_2,\ldots,B_m$;         & \\
         & D :  $D_1\otimes D_2\otimes \ldots\otimes D_k$; \\
          \} \\
\end{tabular}
\end{center}
where $(D_i)_{i\in I}$ is a list of diagram signs, what are smaller than $D$, describing the proper sequence of gluing to form $D$. Note what, every view can be seen as a diagram with only one multi-arrow.
The commutativity or $\lambda$-commutativity for a diagram $D$, for view $G$, in the model, is denote, respectively, by $G:[D]$ or $G:[D]_\lambda$.  When the interpretation for a relation or view $G$ must satisfy a limit or a colimit, we writing
$_{G:lim\;D,\;\; G:\lambda-lim\;D,\;\; G:colim\;D,}$ or $_{G:\lambda-colim\;D.}$ In specification:
 \begin{center}
\tiny
\begin{tabular}{rll}
  $G:\{$ &$A_1, A_2, \ldots, A_n\rightharpoonup B_1, B_2, \ldots, B_m$;         & \\
          &\begin{tabular}{rll}
  $D:\{$ &$A_1,A_2,\ldots,A_n\rightharpoonup  B_1,B_2,\ldots,B_m$;         & \\
         & D :  $D_1\otimes D_2\otimes \ldots\otimes D_k$; & \\\} \\
\end{tabular}&\\
         & $G: \lambda_0$-$lim\;D$;&\\
         &\begin{tabular}{rll}
  $D':\{$ &$A_1,A_2,\ldots,A_n\rightharpoonup  B_1,B_2,\ldots,B_m$;         & \\
         & $D' :  D'_1\otimes D'_2\otimes \ldots\otimes D'_k$; & \\\} \\
\end{tabular}&\\
         & G:$[D']_{\lambda_1}$;&\\
         \} \\
\end{tabular}
\end{center}
 every interpretation for view $G$ must be $\lambda_0$-similar to $Lim\;D$, there $D$ is diagram defined gluing diagrams or multi-arrows $(D_i)_{i\in I}$, and must transform $D'$ is a $\lambda_0$-commutative diagram. However, this type of properties are very generic and some times difficult to understand. When we want to be more specific, we describe some of the relations or views properties using first-order formulas.  We will express internal properties on a relation or view $G$  using formulas, with the same arity as $G$, and defined using only signs that are smaller than $G$ in the library of component. For it we write,
\begin{center}
\tiny
\begin{tabular}{rll}
  $G:=\{$ &$A_1,A_2,\ldots,A_n$;         & \\
         & $G(x_1,x_2,\ldots,x_n):P(R_1(x_1,x_2,\ldots,x_n),R_1(x_1,x_2,\ldots,x_n),\ldots,R_m(x_1,x_2,\ldots,x_n))$; \\
          \} \\
\end{tabular}
\end{center}
if view or relation $G$ interpretation satisfies formula $P$ having the some arity as $R$ and, dependente from relations $R_1,R_2,\ldots,R_n$ that are smaller than $R$ in the library of componentes.

We simplify the specification using some special meta-signs, following the spirit of M. Makkai \cite{Makkai93} and Z. Diskin \cite{Diskin99}. If a view $D$ is a \emph{is\_a} relation  or a clustering relation using the similarity relation $\Gamma$, we write:
\begin{center}
\tiny
\begin{tabular}{rll}
  $D:\{$ &$A_1,A_2,\ldots,A_n\rightharpoonup  B$;         & \\
        &\begin{tabular}{rll}
             $\Gamma:\{$ &$A_1,A_2,\ldots,A_n$;         & \\
                & $\Gamma$ : similarity;\\
          \} \\
\end{tabular}\\
         & $D$ : is\_a$(\Gamma)$; \\
          \} \\
\end{tabular}
\end{center}

In the next section we describe a process useful for knowledge extraction. We are particularly interest in symbolic representation to simplify knowledge integration via relational specifications. There are many methodologists available for this task, however our method uses the same logic assumed to govern the UoD.

\section{\uppercase{Extracting knowledge from concept descriptions}}

Given a concept description ${R:O\rightharpoonup\coprod_{\alpha\in Att}A_\alpha},
$ our goal is the extraction of knowledge from one of its views ${R_{V,U}:O\times\prod_{\alpha\in V}A_\alpha\rightharpoonup\coprod_{\alpha\in U}A_\alpha}$. For that the information structure is crystallized in a neural network and codified in string-based notation as a formula. Different concept can be seen as answers to different queries to a UoD models. In this sense different concepts represent different perspective for the available data, allowing the enrichment of a knowledge base or specification systems with new insights. In this section, we present a methodology to extract first-order formulas using neural networks describing  available information in a {\L}logic.

As mentioned in \cite{Amato02} there is a lack of a deep investigation of the relationships between logics and NNs. In \cite{Castro98} it is shown how, by taking as activation function, $\psi$, the identity truncated to zero and one,
\[
_{\psi(x)=\min(1,\max(x, 0)),}
\]
it is possible to represent the corresponding NN as a combination of propositions of {\L}ukasiewicz calculus and \emph{vice-versa} \cite{Amato02}.

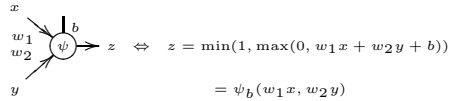
\begin{figure}[h]
\begin{center}
\tiny
$
\xymatrix @R=5pt @C=8pt { x\ar[dr]_{w_1} & \ar@{-}[d]^{b} & \\
                           & *+[o][F-]{\psi} \ar[r]&z\;\;\Leftrightarrow\;\; z=\min(1,\max(0,w_1x+w_2y+b)) \\
           y\ar[ur]^{w_2} &  & \;\;=\psi_b(w_1x,w_2y) \
           }
$
\end{center}
\caption{functional interpretation for a NN }\label{interpretation}
\end{figure}

For used NNs to learn {\L}ukasiewicz sentences, we define the first-order language  as a set of circuits generated from the plugging of atomic components. For this, we used the library of components presented in table \ref{semioticaLuk}, interpreted as neural units and we gluing them together, to form NNs having only one output, without loops. This task of construct complex structures based on simplest ones can be formalized using generalized programming \cite{Fiadeiro97}.
\begin{table}
\begin{center}
\tiny
\begin{tabular}{|c|c||c|c||c|c||c|c|}
  \hline
  Formula: & Configuration: & Formula: & Configuration: & Formula: & Configuration: & Formula: & Configuration: \\
  \hline
  \hline
  $\neg x\oplus y$
  &
  $\xymatrix @R=6pt @C=6pt { x\ar[dr]_{-1} & \ar@{-}[d]^{1} & \\
                           & *+[o][F-]{\varphi} \ar[r]& \\
           y\ar[ur]^{1} &  &  \
           }$
  &
  $ x\otimes \neg y$
  &
  $\xymatrix @R=6pt @C=6pt { x\ar[dr]_{1} & \ar@{-}[d]^{0} & \\
                           & *+[o][F-]{\varphi} \ar[r]& \\
           y\ar[ur]^{-1} &  &  \
           }$
  &
  $x\oplus y$
  &
  $\xymatrix @R=6pt @C=6pt { x\ar[dr]_{1} & \ar@{-}[d]^{0} & \\
                           & *+[o][F-]{\varphi} \ar[r]& \\
           y\ar[ur]^{1} &  &  \
           }$
  &
  $\neg x\otimes \neg y$
  &
  $\xymatrix @R=6pt @C=6pt { x\ar[dr]_{-1} & \ar@{-}[d]^{1} & \\
                          & *+[o][F-]{\varphi} \ar[r]& \\
           y\ar[ur]^{-1} &  &  \
           }$ \\
  \hline
  $x\oplus \neg y$ & $\xymatrix @R=6pt @C=6pt { x\ar[dr]_{1} & \ar@{-}[d]^{1} & \\
                           & *+[o][F-]{\varphi} \ar[r]& \\
           y\ar[ur]^{-1} &  &  \
           }$ & $x\otimes y$ & $\xymatrix @R=6pt @C=6pt { x\ar[dr]_{1} & \ar@{-}[d]^{-1} & \\
                           & *+[o][F-]{\varphi} \ar[r]& \\
           y\ar[ur]^{1} &  &  \
           }$ &
  $\neg x\otimes y$ & $\xymatrix @R=6pt @C=6pt { x\ar[dr]_{-1} & \ar@{-}[d]^{0} & \\
                           & *+[o][F-]{\varphi} \ar[r]& \\
           y\ar[ur]^{1} &  &  \
           }$ &
           $\neg x \oplus \neg y$
             &
             $\xymatrix @R=6pt @C=6pt { x\ar[dr]_{-1} & \ar@{-}[d]^{2} & \\
                           & *+[o][F-]{\varphi} \ar[r]& \\
           y\ar[ur]^{-1} &  &  \
           }$
               \\
           \hline
\end{tabular}
\end{center}
\caption{Possible configurations for a neuron in a {\L}NN a its interpretation.}\label{semioticaLuk}
\end{table}
The neurons of these types of networks, which have two inputs and one output, can be interpreted as a function (see figure \ref{interpretation}) and are generically denoted, in the following, by $\psi_b(w_1x_1,w_2x_2)$, where $b$ represent the node bias, $w_1$ and $w_3$ are the weights and, $x_1$ and $x_2$ input values. We simplify exposition by calling to $w_1x_1$ and $w_2x_2$ \emph{variables} in $\psi_b$. In this context a network is the \emph{functional interpretation} of a formula in the string-based notation when the relation, defined by network execution, corresponds to the formula truth table.
The use of NNs as interpretation of formulas simplifies the transformation between string-based representations and the network representation, allowing one to write:

\begin{prop}\label{prop1}
Every well-formed formula in the {\L}logic language can be codified using a NN, and the network defines the formula interpretation, when the activation function is the identity truncated to zero and one.
\end{prop}

For instance, the semantic for sentence $_{\varphi=(x\otimes y\Rightarrow z)\oplus(z \Rightarrow w),}$ can be described using the bellow network or can be codified by the presented set of matrices. From this matrices we must note that the local interpretation of each unit is a simple exercise of pattern checking, where we take by reference the existent relation between formulas and configuration described in table \ref{semioticaLuk}.
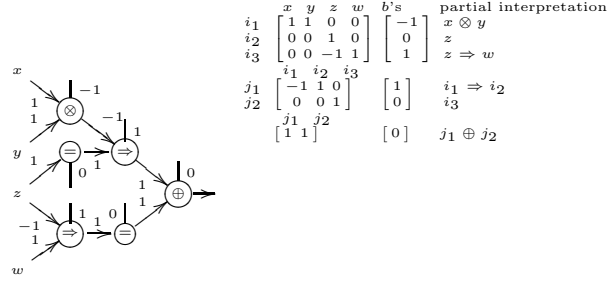
\begin{center}
\tiny
$
\xymatrix @R=6pt @C=8pt { x\ar[dr]_{1} & \ar@{-}[d]^{-1} & \\
                           & *+[o][F-]{\otimes} \ar[rd]^{-1}& \ar@{-}[d]^{1} \\
           y\ar[ur]^{1} & *+[o][F-]{=}\ar[r]_{1}  &  *+[o][F-]{\Rightarrow} \ar[rd]_{1} &\ar@{-}[d]^{0}\\
           z\ar[dr]_{-1} \ar[ur]^{1}& \ar@{-}[d]^{1}\ar@{-}[u]_{0} & \ar@{-}[d]_{0} &*+[o][F-]{\oplus} \ar[r]&\\
                           & *+[o][F-]{\Rightarrow} \ar[r]^{1}&*+[o][F-]{=} \ar[ru]^{1}\\
           w\ar[ur]^{1} &  &  \\
           }
$
 \begin{tabular}{llll}
 &$\begin{array}{cccc}
      \;x &\; y &\; z &\; w \\
    \end{array}$ & $b$'s & \tiny partial interpretation
      \\
$ \begin{array}{c}
      i_1  \\
      i_2 \\
      i_3 \\
    \end{array}$
    &
  $\left[
    \begin{array}{cccc}
      1 &  1 & 0 & 0 \\
      0 &  0 & 1 & 0 \\
      0 &  0 & -1 & 1\\
    \end{array}
  \right]
  $ &$ \left[
       \begin{array}{c}
         -1 \\
         0 \\
         1 \\
       \end{array}
     \right]
    $& $\begin{array}{l}
         x \otimes y \\
         z  \\
         z\Rightarrow w\\
       \end{array}$\\
    &$\begin{array}{ccc}
      \;i_1 & \;i_2 & \;i_3 \\
    \end{array}$& &
       \\
       $
  \begin{array}{c}
      j_1 \\
      j_2 \\
    \end{array}$
  &
  $\left[
    \begin{array}{ccc}
      -1 &  1 & 0 \\
       0 &  0 & 1 \\
    \end{array}
  \right]
   $&$ \left[
       \begin{array}{c}
         1 \\
         0 \\
       \end{array}
     \right]
    $& $\begin{array}{l}
         i_1\Rightarrow i_2 \\
        i_3 \\
       \end{array}$\\
   &$\begin{array}{cc}
         \;j_1 &\;j_2 \\
    \end{array}$
       \\
  &$\left[
      \begin{array}{cc}
         1 & 1 \\
       \end{array}
     \right] $& $\left[
       \begin{array}{c}
        0 \\
       \end{array}
     \right]$ & $j_1\oplus j_2$\\
\end{tabular}
\end{center}
\begin{center}
\tiny
\begin{tabular}{lll}
   INTERPRETATION:&\\
   &$j_1\oplus j_2=(i_1\Rightarrow i_2)\oplus (i_3)=((x \otimes y)\Rightarrow z)\oplus (z\Rightarrow w)$\\
\end{tabular}
\end{center}
In this sense this NN can be seen as an interpretation for sentence $\varphi$; it codifies $f_\varphi$, the proposition truth table.
\[
_{f_\varphi(x,y,z,w)=\psi_0(\psi_0(\psi_1(-z,w)),\psi_1(\psi_0(z),-\psi_{-1}(x,y)))}
\]
However, truth table $f_\varphi$ is a continuous structure, for our goal, it must be discretized and represented using a finite structure, ensuring sufficient information to describe the original formula. A truth table $f_\varphi$ for a formula $\varphi$, in a fuzzy logic, is a map $f_\varphi:[0,1]^m\rightarrow [0,1]$, where $m$ is the number of propositional variables used in $\varphi$. For each integer $n>0$, let $S_n$ be the set $\{0,\frac{1}{n},\ldots,\frac{n-1}{n},1\}$. Each $n>0$, defines a sub-table for $f_\varphi$ defined by $f_\varphi^{(n)}:(S_n)^m\rightarrow [0,1]$, given by $f_\varphi^{(n)}(\bar{v})=f_\varphi(\bar{v})$, and called the $\varphi$ \emph{(n+1)-valued truth sub-table}.

\subsection{Similarity between a configuration and a formula}

We call \emph{Castro neural network} (CNN) a type of NN having as activation function $\psi(x)=\min(1,max(0,x))$, where its weights are -1, 0 or 1 and having by bias an integer. A CNN is called \emph{{\L}ukasiewicz neural network} ({\L}NN) if it can be codified as a binary NN: i.e. a  CNN where each neuron has one or two inputs. A CNN is \emph{representable} when can be codified using an equivalent {\L}NN.

Each neuron, with $n$ inputs, in a CNN can be described using configuration $$_{\alpha=\psi_b(x_1,x_2,\ldots,x_{n-1},x_n)}$$ and it is representable when can be describes by a {\L}NN  $$_{\alpha=\psi_{b_1}y_1,\psi_{b_2}(y_2,\psi_{b_3}(\ldots,\psi_{b_{n-1}}(y_{n-1},y_n))).}$$
A CNN is representable if each of its neurons is representable. Note that, a representable CNN  can be translated directly into {\L}ukasiewicz first-order language, using the correspondences between configurations and formulas described on table \ref{semioticaLuk}.

Given the configuration $_{\alpha=\psi_b(x_1,x_2,\ldots,x_n)}$, in a CNN,  with $_{0\leq x_1+x_2+\ldots+x_n+b\leq 1}$, we have $_{0\leq x_1+(x_2+\ldots+x_n+b_2)+b_1\leq 1}$, where $_{b=b_1+b_2}$ for integers $b_0$ and $b_1$. And we have $$_{\psi_b(x_1,x_2,\ldots,x_n)=\psi_{b_1}(x_1,\psi_{b_2}(x_2,\ldots,x_n))}.$$

Naturally, a neuron configuration - when representable - can by codified by
different {\L}NN. Particularly, we have:
\begin{prop}
If the neuron configuration $_{\alpha=\psi_b(x_1,x_2,\ldots,x_{n-1},x_n)}$ is representable, but not constant, it can be codified in a {\L}NN  with the following structure:\\
$$
_{\alpha=\psi_{b_1}(x_1,\psi_{b_2}(x_2,\ldots,\psi_{b_{n-1}}(x_{n-1},x_n)\ldots)),}
$$
where $_{b_1,b_2,...,b_{n-1}}$ are integers, and  $_{b=b_1+b_2+...+b_{n-1}}$.
\end{prop}

And, since the $n$-nary operator $\psi_b$ is commutative, variables $_{x_1,x_2,\ldots,x_{n-1},x_n)}$ could interchange its position in function $_{\alpha=\psi_b(x_1,x_2,\ldots,x_{n-1},x_n)}$ without changing the operator output. By this we mean that, for a three input configuration, when we permutate variables, we  generate equivalent configurations:
$$_{\psi_b(x_1,x_2,x_3)=\psi_b(x_2,x_3,x_1)=\psi_b(x_3,x_2,x_1)=\ldots}$$
When these are representable, they can be codified in string-based notation using logic connectives. But these different configuration only generate equivalent formulas if these formulas are  disjunctive or conjunctive. A disjunctive formulas is formula written using the disjunction of propositional variables or negation of propositional variable. Similarly, a  conjunctive formulas are formulas written using only the conjunction of propositional variables or its negation.

\begin{prop}
If $_{\alpha=\psi_b(x_1,x_2,\ldots,x_{n-1},x_n)}$ is representable, it is the interpretation of a disjunctive formula or a conjunctive formula.
\end{prop}

This leave us with the task of classifying a neuron configuration according to its representation. For that, we must note what, if
$$
_{\alpha=\psi_b(-x_1,-x_2,\ldots,-x_n, x_{n+1},\ldots,x_m)}
$$
is representable:
\begin{enumerate}
\item When $b=n$ is the number of negative inputs, in $\alpha$, we have
$$
_{\alpha=\psi_{1}(-x_1,\psi_{1}(-x_2,\ldots\psi_{1}(-x_{n}, \psi_{0}(x_{n+1},\ldots\psi_{0}(x_{m-1},x_m))\ldots)\ldots)),}
$$
using Table \ref{semioticaLuk}, the configuration $\alpha$ is the interpretation for
$$
_{\neg x_1\oplus\ldots\oplus\neg x_n\oplus x_{n+1}\oplus\ldots\oplus x_m.}
$$

\item When $b=-p+1$ is the number of negative inputs, in $\alpha$, we have
$$
_{\alpha=\psi_{1}(-x_1,\psi_{1}(-x_2,\ldots\psi_{0}(-x_{n}, \psi_{-1}(x_{n+1},\ldots\psi_{-1}(x_{m-1},x_m))\ldots)\ldots)),}
$$
an interpretation for formula
$$
_{\neg x_1\otimes\ldots\otimes\neg x_n\otimes x_{n+1}\otimes\ldots\otimes x_m.}
$$
\end{enumerate}

This establishes a relationship between the formula structure and the configuration bias, the number of negative and positive weights.
\begin{prop}\label{conf classification}
Given the neuron configuration
$
_{\alpha=\psi_b(-x_1,-x_2,\ldots,-x_n, x_{n+1},\ldots,x_m)}
$
with $m=n+p$ inputs and where $n$ and $p$ are, respectively, the number of negative and the number of positive weights, on the neuron configuration:
\begin{enumerate}
  \item If $b=-p+1$ the neuron is called a \emph{conjunction} and it is an interpretation for
$
_{\neg x_1\otimes\ldots\otimes\neg x_n\otimes x_{n+1}\otimes\ldots\otimes x_m.}
$
  \item When $b=n$ the neuron is called a \emph{disjunction} and it is an interpretation of
$
_{\neg x_1\oplus\ldots\oplus\neg x_n\oplus x_{n+1}\oplus\ldots\oplus x_m.}
$
\end{enumerate}
\end{prop}

Imposing some structural order on the neural network transformation:

\begin{prop}
Every conjunctive or disjunctive configuration \\ $$_{\alpha=\psi_b(x_1,x_2,\ldots,x_{n-1},x_n),}$$ can be codified by a  {\L}NN \\ $$_{\beta=\psi_{b_1}(x_1,\psi_{b_2}(x_2,\ldots,\psi_{b_{n-1}}(x_{n-1},x_n)\ldots)),}$$
where $$_{b_1,b_2,...,b_{n-1}\text{ are integers, }b=b_1+b_2+\cdots+b_{n-1}\text{ and }b_1\leq b_2\leq \cdots\leq b_{n-1}.}$$
\end{prop}

This property can be translated in the following rewriting rule,
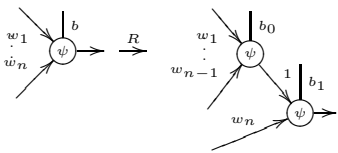
\[
\tiny
\xymatrix @R=6pt @C=10pt {  \ar[rd]_{w_1} & \ar@{-}[d]^{b} &    &  \\
                      \vdots     & *+[o][F-]{\psi} \ar[r] & \ar[r]^{R} &\\
            \ar[ru]^{w_n} & &\\
           }
\xymatrix @R=7pt @C=8pt {  \ar[rd]_{w_1} & \ar@{-}[d]^{b_0} &    &  \\
                   \vdots        & *+[o][F-]{\psi} \ar[rd]^{1} & \ar@{-}[d]^{b_1}\\
            \ar[ru]^{w_{n-1}} & &*+[o][F-]{\psi}\ar[r]&\\
            \ar[rru]^{w_{n}} & &\\
           }
\]
linking equivalent networks, when the integers $b_0$ and $b_1$ satisfy $b=b_0+b_1$ and $b_1\leq b_0$, and are such that neither of the involved neurons have constant output. Note that, a representable CNN can be transformed by the application of rule R in a set of equivalent {\L}NN with simplest neuron configuration:
\begin{prop}
Un-representable neuron configurations are those transformed by
rule R in, at least, two non-equivalent NNs.
\end{prop}

For instance, the un-representable configuration $\psi_0(-x_1,x_2,x_3)$, is transformed by rule R in three non-equivalent configurations:
\begin{center}
\tiny
\begin{tabular}{|c|c|}
  \hline
  & \\
  $\psi_0(x_3,\psi_0(-x_1,x_2))=f_{x_3\oplus(\neg x_1\otimes x_2)}$ & $\psi_{-1}(x_3,\psi_{1}(-x,x_2))=f_{x_3\otimes(\neg x_1\otimes x_2)}$ \\ $\psi_0(-x_1,\psi_0(x_2,x_3))=f_{\neg x_1\otimes(x_2\oplus x_3)}$ &\\
  & \\
  \hline
\end{tabular}
\end{center}
The representable configuration $\psi_2(-x_1,-x_2,x_3)$ is transformed by rule R on only two distinct but equivalent configurations:
\begin{center}
\tiny
\begin{tabular}{|c|c|}
  \hline
  & \\
  $\psi_0(x_3,\psi_2(-x_1,-x_2))=f_{x_3\oplus \neg (x_1\otimes x_2)}$ & $\psi_1(-x_2,\psi_1(-x_1,x_3))=f_{\neg x_2\oplus (\neg x_1\oplus x_3)}$ \\
  & \\
  \hline
\end{tabular}
\end{center}

For the extraction of knowledge from trained NNs, we  translate neuron configuration in propositional connectives to form formulas. However, not all neuron configurations can be translated in formulas, but they can be approximate by one. To quantify the approximation quality we used the exponential-similarity.

Two neuron configurations $\alpha=\psi_{b}(x_1,x_2,\ldots,x_n)$ and $\beta=\psi_{b'}(y_1,y_2,\ldots,y_n)$, are  called $\lambda$-similar,  in a $(m+1)$-valued {\L}logic, if  ${\lambda=e^{-\frac{1}{|O|}\sum_{\bar{x}\in T}\neg(\alpha(\bar{x})\Leftrightarrow\beta(\bar{x}))}}$, we write $
\alpha\sim_\lambda\beta$.
If $\alpha$ is un-representable and $\beta$ is representable, the second configuration is called \emph{a representable approximation} to the first.

On the $2$-valued {\L}logic (the Boolean logic case), we have for the un-representable configuration $\alpha=\psi_0(-x_1,x_2,x_3)$:
\begin{center}
\tiny
\begin{tabular}{|c|c|}
  \hline
  & \\
  $\psi_0(-x_1,x_2,x_3)\sim_{0.883}\psi_0(x_3,\psi_0(-x_1,x_2))$& $\psi_0(-x_1,x_2,x_3)\sim_{0.883}\psi_{-1}(x_3,\psi_{1}(-x_1,x_2))$\\
  $\psi_0(-x_1,x_2,x_3)\sim_{0.883}\psi_0(-x_1,\psi_0(x_2,x_3))$& \\
  & \\
  \hline
\end{tabular}
\end{center}
In this case, the truth sub-tables of, formulas $\alpha_1=x_3\oplus(\neg x_1\otimes x_2)$, $\alpha_1=x_3\otimes(\neg x_1\otimes x_2)$ and $\alpha_1=\neg x_1\otimes(x_2\oplus x_3)$ are both $\lambda$-similar to $\psi_0(-x_1,x_2,x_3)$, where $\lambda=0.883$, since they differ in one position on 8 possible positions. This means that both formulas are 92\% accurate.

For an un-representable configuration, $\alpha$, we can generate the finite set $S(\alpha)$, of representable networks similar to $\alpha$, using rule R. Given a $(n+1)$-valued logic, from that set of formulas we select to approximate $\alpha$ the formula having the interpretation more similar to $\alpha$. This identification of un-representable configuration, using representable approximations,  is used to transform networks with un-representable neurons into representable structures. The stress associated with this transformation characterizes the translation quality.

Bellow we present an example of a un-representable CNN:
\[
\tiny
\begin{tabular}{lll}
  $\left[
    \begin{array}{ccccccc}
      -1 &  1 & -1 & 1 & 0 & -1 & 0 \\
      0 &  0 & 0 & 1 &  1 & 0 & -1 \\
      1 &  1 & 0 & 0 &  0 & 0 & -1 \\
    \end{array}
  \right]
  $ &$ \left[
       \begin{array}{c}
         0 \\
         1 \\
         0 \\
       \end{array}
     \right]
    $& $\begin{array}{l}
         i_1\text{ un-representable} \\
         A4\oplus A5\oplus \neg A7  \\
         i_3\text{ un-representable} \\
       \end{array}$\\
  $\left[
    \begin{array}{ccc}
      1 &  -1 & 1 \\
    \end{array}
  \right]
   $&$ \left[
       \begin{array}{c}
         0 \\
       \end{array}
     \right]
    $& $\begin{array}{l}
         j_1 \text{un-representable} \\
       \end{array}$\\
\end{tabular}
\]
For each local un-representable configuration $\alpha$, we selected the most similar representable configuration on $S(\alpha)$, after applying rule R, we have in this case:
\begin{enumerate}
\tiny
  \item $i_1\sim_{0.9387}((\neg A1\otimes A4) \oplus A2)\otimes \neg A3 \otimes \neg A6$,
  \item $i_3\sim_{0.8781}(A1\oplus\neg A7)\otimes A2$, and
  \item $j_1\sim_{0.8781}(i_1\otimes\neg i_2)\oplus i_3$.
\end{enumerate}
Using this substitutions we reconstructed the formula:
\begin{center}
\tiny
$\alpha=(((((\neg A1\otimes A4) \oplus A2)\otimes \neg A3 \otimes \neg A6)\otimes\neg (A4\oplus A5\oplus \neg A7))\oplus ((A1\oplus\neg A7)\otimes A2)$,
\end{center}
$\lambda$-similar to the original CNN, with $\lambda=0.7323$, in a $5$-valued {\L}logic.

\subsection{Crystallizing trained neural networks}

Standard error back-propagation algorithm (EBP) is a gradient descent algorithm, in which the network weights are moved along the negative of the gradient of the performance function. EBP algorithm has been a significant improvement in NN research, but it has a weak convergence rate. Many efforts have been made to speed up the EBP algorithm. The Levenberg-Marquardt (LM) algorithm \cite{HaganMenhaj99} \cite{Andersen95} ensued from the development of EBP algorithm-dependent methods. It gives a good exchange between the speed of the Newton algorithm and the stability of the steepest descent method \cite{Battiti92}.

The basic EBP algorithm adjusts the weights in the steepest descent direction.
When training with the EBP method, an iteration of the algorithm defines the change of weights and has the form
$
_{w_{k+1}=w_k-\alpha G_k,}
$
where $G_k$ is the gradient of performance index $F$ on $w_k$, and $\alpha$ is the learning rate.

Note that, the basic step of Newton's method can be derived from Taylor formula and is $
_{w_{k+1}=w_k-H_k^{-1}G_k,}
$
where $H_k$ is the Hessian matrix of the performance index at the current values of the weights.

Since Newton's method implicitly uses quadratic assumptions, the Hessian matrix dos not need be evaluated exactly. Rather, an approximation can be used, such as
$
_{H_k\approx J_k^TJ_k,}
$
where $J_k$ is the Jacobian matrix that contains first derivatives of the network errors with respect to the weights $w_k$.

The simple gradient descent and Newtonian iteration are complementary in the advantages they provide. Levenberg proposed an algorithm based on this observation, whose update rule blends aforementioned algorithms and is given as
\[
_{w_{k+1}=w_k-[J_k^TJ_k+\mu I]^{-1}J_k^Te_k},
\]
where $e_k$ is a vector of current network errors and $\mu$ is the learning rate. This update rule is used as follows. If the error goes down following an update, it implies that our quadratic assumption on the function is working and we reduce $\mu$ (usually by a factor of 10) to reduce the influence of gradient descent. In this way, the performance function is always reduced at each iteration of the algorithm \cite{Megan96}. On the other hand, if the error goes up, we would like to follow the gradient more and so $\mu$ is increased by the same factor.

We can obtain some advantage out of the second derivative, by scaling each component of the gradient according to the curvature. This should result in larger movements along the direction where the gradient is smaller so the classic "error valley" problem does not occur any more. This crucial insight was provided by Marquardt. He replaced the identity matrix in the Levenberg update rule with the diagonal of Hessian matrix approximation resulting in the  LM update rule.

In each LM iteration we restricted the NN representation bias, making  its structure similar to a CNN. For that, we used a \emph{smooth crystallization} procedure resulting from function,
 \[
 _{\Upsilon_n(w)=sign(w).((\cos(1-abs(w)-\lfloor abs(w)\rfloor).\frac{\pi}{2})^n+\lfloor abs(w)\rfloor),}
 \]
iteration, where $sign(w)$ is the sign of $w$ and $abs(w)$ its absolute value. Denoting by $\Upsilon_n(N)$ the function having by input and output a NN, where the weights on the output network results of applying  $\Upsilon$ to all the input network weights and neurons biases. Each interactive application of $\Upsilon$ produce a networks progressively more similar to a CNNs.  For show that, we define by \emph{representation error}, for a network $N$, with weights $w_1,\ldots,w_n$,
 \[
 \Delta(N)=\sum^n_{i=1}(w_i-\lfloor w_i\rfloor).
 \]
 When $N$ is a CNNs we have $\Delta(N)=0$. Since, for every network $N$ and $n>0$, $\Delta(N)\geq \Delta(\Upsilon_n(N))$, we have
\begin{prop} Given a neural networks $N$ with weights in the interval $[0,1]$. For every $n>0$ the function $\Upsilon_n(N)$ have by fixed points {\L}ukasiewicz neural networks $N'$.
\end{prop}

We changed the LM  algorithm by applying a soft crystallization step after the  LM update rule:
\[
_{w_{k+1}=\Upsilon_2(w_k-[J_k^TJ_k+\mu
.diag(J_k^TJ_k)]^{-1}J_k^Te_k})
\]
This can be seen as a process to regularize the network and improves drastically the convergence to a CNN preserving its ability to learn. On our method network regularization is made using three different strategies:
\begin{enumerate}
  \item using the described soft crystallization process, where we restricted the knowledge dissemination on the network structure, information is concentrated on some weights;
  \item after the training we use crisp crystallization, where links between neurons with weights near 0 are removed and weights near -1 or 1 are consolidated;
  \item the resulting crisp network is pruned using ''\emph{Optimal Brain Surgeon}'' method.
\end{enumerate}
The first regularization technic avoids knowledge dissemination on the NN. The last regularization technic avoid redundancies, in the sense that the same or redundant information can be codified at different locations. We minimized this by selecting weights to eliminate. The ''\emph{Optimal Brain Surgeon}'' method uses the criterion of minimal increase in training error. It uses information from all second-order derivatives of the error function to perform network pruning.

The \emph{Optimal Brain Surgeon} method is derived from  Taylor series of the error with respect to weights,
\[
\Delta E=J_w^T.\Delta w + \frac{1}{2}\Delta w^TH_w\Delta w+O(\|\Delta w\|^3).
\]
For a network trained to a local minimum in error, $\Delta w\approx0$, the first linear term vanishes, third and all higher order terms are ignored. The method finds a weight to be set to zero (which we call $w_q$) to minimize $\Delta E$ the increase in error. Making $w_q$ zero correspond to change its value using $\Delta w_q$ making $\Delta w_q+w_q=0$ or more generally:
\[
e^T_q\Delta w+w_q=0,
\]
where $e_q$ is the unit vector in weight space corresponding to (scalar) weight $w_q$. This reduces our goal to solve:
\[
\min_q\{\min_{\Delta w}\frac{1}{2}\Delta w^TH_w\Delta w\;|\;e^T_q\Delta w_q+w_q=0\}
\]
The optimization problem is constrained, following \cite{Hassibi93}, we form the Lagrangian operator
\[
L=\frac{1}{2}\Delta w^TH_w\Delta w+\lambda(e^T_q\Delta w_q+w_q),
\]
where $\lambda$ is a Lagrange undetermined multiplier. After take functional derivatives, employ the constraint $e^T_q\Delta w+w_q=0$, and use matrix inverse to find that the optimal weight change and resulting change in error are respectively
\[
\Delta w= - \frac{w_q}{[H_w^{-1}]_{qq}}H_w^{-1}e_q\text{ and }
L_q=\frac{1}{2}\frac{w^2_q}{[H_w^{-1}]_{qq}}.
\]
The method recalculates the magnitude of all the weights in the NN. Hassibi,  Stork and Stork called $L_q$ the "saliency" of weight $q$, the increase in error that results when the weight is eliminated.

Algorithm \ref{RevEng} describes our methodology for training CNN and extraction of symbolic pattern descriptions.
\begin{algorithm}
\tiny
\caption{Reverse Engineering algorithm} \label{RevEng}
\begin{algorithmic}[1]
\STATE Given a concept description evaluated on a (n+1)-valued  {\L}ukasiewicz logic
\STATE Define an initial network complexity
\STATE Generate an initial neural network
\STATE Apply the LM algorithm with soft crystallization
\IF{ the generated network have bad performance}
\STATE If need increase network complexity
\STATE Try a new network. Go to 3
\ENDIF
\STATE Crisp crystallization on the trained NN.
\IF{crystallized network have bad performance}
\STATE Try a new network. Go to 3
\ENDIF
\STATE Refine the crystallized NN using ''\emph{Optimal Brain Surgeon}'' algorithm
\STATE Identify un-representable configurations
\STATE Replace each un-representable configurations, using a similar representable configuration, selected from the set of configurations generated using rule $R$.
\STATE Evaluated the procured NN performance on the original concept description.
\STATE Translated the procured NN on string-based notation.
\end{algorithmic}
\end{algorithm}

Given a view for a concept description, we try to find a CNN describing information. For that our implementation generates neural networks with a fixed number of hidden layers (in our implementation we used three). When a network have bad learning performance, training is stopped, and train is initiate for a new network, with random heights. After a fixed number of tries the network topology is changed. This number iterations depends on the number of network inputs. After trying configure a set of networks with a given complexity and bad learning performance, the system tries to apply LM algorithm, with soft crystallization, for a more complex set of networks. The process stops when a adequate description for data is find. And after the network be pruned, un-representable configurations are approximated using representable ones. This defines a description for the information using a {\L}NN.

\section{\uppercase{Generating artificial fuzzy structures}}

We tested our methodology in real data sets \cite{Leandro09} and on artificially generated data structures. On this section we present our approach using artificially generated data sets. For that, we developed a simple way to generate complex fuzzy structures. Our method is based on a non-deterministic state machine. In this machine every states have associated a level of uncertainty quantified by a truth value, from a many-valued logic $\Omega$. These type of machines have its stats changed based on the reading a word and this change is described by a relation defining stat transition, from the actual state and the signs that are being read. This relation, for stat change, is a multi-morphism evaluated in $\Omega$ and the actual signs being read are described by a $\Omega$-set. We called to these type of state machines a $\Omega$-\emph{automata}.

In this sections we describe an automata as a concept to be learned, and show how reverse engineering its structure using the data generated from its execution. We do this translating a CNN structure in a formula, representable on a specification system. Beginning with a structural description of a problem, in the from of a specification system, and a model, we enrich the specification structure using knowledge extracted from data generated querying the model and crystallized on neuronal networks.

\subsection{{\L}ukasiewicz automatas}

A $\Omega$-automata is an non-deterministic state machine, where state transition relation. This change is made after reading a sign from a word. Each symbol used to define automata input string is a sign having the form $\alpha=_{\lambda}x$, where $\alpha\in Att$ is an attribute, $x$ a possible value for $\alpha$, $x\in A_\alpha$, and $\lambda$ quantifies the truth value of $\alpha=x$ evaluated in $\Omega$. The input sting at position $i$ defines the truth of $\alpha=_{\lambda}x$  for each $\alpha\in Att$. The automata is defined by a set of states $E$, and at each moment this states have associated a level of uncertainty in $\Omega$ of being the automata state. In this sense an $\Omega$-automata actual state is describes by a relation evaluated in $\Omega$, $e_i:E\rightarrow \Omega$, having by support the set of automata possible states. The states of an automata are classified in three classes: input states, auxiliary states and output states. The input states have its uncertainty directly assigned by the uncertainty on the input signs. Each of this state is associated directly with one sign $\alpha= x$, used on the input string, and its uncertainty is the truth value of equality $\alpha= x$ on the actual reading position.  After all input string have been read, the level of uncertainty on output states defines the $\Omega$-automata output. An $\Omega$-automata begins its activity in a initial state $e_0:E\rightarrow \Omega$, reads a string $s_1s_2\ldots s_n$, in each iteration a position $s_i$ is read and on the $n$-iteration reach its final state $e_n:E\rightarrow \Omega$. The final state is used to describe the automata output $o:E_O\rightarrow \Omega$. Formally

\begin{defn}
 An $\Omega$-automata $A$ is a structure described by  $$(\text{\L},(A_i)_{i\in Att},E,E_O,\{M_\lambda,M_{\neg\lambda}\}_{\lambda\in\Sigma},e_0)$$
where:
\begin{enumerate}
  \item {\L} is a finite multi-valued logic having truth values in $\Omega$;
  \item $(A_\alpha)_{\alpha\in Att}$ a family $\Omega$-sets used as domain for attributes in construction of signs $'\alpha=x'$;
  \item $E$ is the set of states, where each input sign $'\alpha=x'$ have a state associated. The states, with a sign associated, are called \emph{input states};
  \item $E_O$ is a subset of $E$, called set \emph{output states};
  \item two boolean matrices $M_0$ and $M_1$, describing state transition. $M_0$ describes negative uncertainty propagation and $M_1$ positive uncertainty propagation;
  \item $e_0:E\rightarrow \Omega$ is a relation describing the initial automata state;
  \item if the automata state on iteration $k$ is $e_k:E\rightarrow \Omega$ and let $X_k:\coprod_{\alpha \in Att} A_\alpha \rightarrow \Omega$ describe the input sign on position $k$, the new automata state is given by
      \[
       e_{k+1}=M_1(e_k)\oplus M_0(\neg e_k),
      \]
      where in $e_k:E$ we update, the input states, with the sign uncertainty  on the reading position, described by vector $X_k$.
\end{enumerate}
A $\Omega$-automata is called  a \emph{{\L}ukasiewicz automata} when the system is governed by a finite {\L}ukasiewicz logic.
\end{defn}
We named \emph{relational automata} to an extension to the $\Omega$-automata defined using transition matrices $M_0$ and $M_1$ with values in $\Omega$. This type of fuzzy automata have its behaviour described using formulas of Relational {\L}ukasiewicz  logic, introduced in \cite{Gerla01}, outfitting the scope of this work.

In this sense we interpret a  word as a string of begs, where the possibility of a sign is in the beg is described by a relation $X_k:\coprod_{\alpha \in Att} A_\alpha \rightarrow \Omega$. In each iteration the $\Omega$-automata reads a position on the string. If in the iteration $k$ the position $k$ is read, on the position $k+1$ it reads the string position $k+1$. Each position in the string can have more than a sign or can be empty. On iteration $k$ the automata reads the position $k$, updating the uncertainty on each input states, using the sign uncertainties on the reading positions. This uncertainty is propagating to other states, applying the state transformation matrices. The update and the propagation of state uncertainty is done for each iteration. The input string length determines the number of automata iteration.

In this context a word $w$ can be define as a sequence of relations\[
\tiny
s_1, s_2,\ldots s_i\ldots,s_n
\]
having the type $s_i:\coprod_{\alpha\in Att}A_\alpha\rightarrow \Omega$. Using $w$ as input to the automata, it
generates a sequence of states described using a sequence of relation
\[
\tiny
e_1,e_2,\ldots e_i\ldots,e_{n},
\] having the type $e_i:E\rightarrow \Omega$, describing the change on automata state between iterations. The $\Omega$-set $e_1$ is defined using the automata initial state $e_0$, where the input states were update with the reading position $s_0$. The state $e_i$, when $i>1$, depends on  state $e_{i-1}$ updated with the uncertainty  on input sign, described by the reading position  $s_i$. The automata output is defined by the uncertainty in each output state $E_O$ in the automata state $e_{n}$. The following example illustrates an automata execution.

\begin{exam}[A binary {\L}ukasiewicz automata]\label{exemautbin}
 The string of a binary {\L}ukasiewicz automata is defined using only an attribute, having two possible values. Let this attribute be $a$ and its possible values $0$ or $1$. Words interpreted using this automata are described using a sequence of $\Omega$-sets having by support the set of signs $\{'a=0','a=1'\}$. An example, of this type of words is presented on table \ref{word}, where each column codifies the existence of each sign in that position.
\begin{table}
\begin{center}
\small
 \begin{tabular}{|r|llllllllllll|}
   \hline
   '$a$=1' & 1 & 1 & 1/2 & 0 & 1/4 & 1/2 & 1 & 1 & 1/2 & 1/4 & 0 & 0 \\
   '$a$=0' & 0 & 0 & 1/2 & 1 & 3/4 & 1/2 & 0 & 0 & 1/2 & 3/4 & 0 & 1 \\
   \hline
 \end{tabular}
 \end{center}
 \caption{Word defined using signs '$a$=1' and '$a$=0'.}\label{word}
\end{table}
For this example, we fixed the finite {\L}ukasiewicz logic having by truth values $_{\Omega=\{0,1/4,1/2,3/4,1\}}$. This table can be  interpreted by saying what: first and second marks in the word are a '1', position 4 and 12 have a '0', and in position 11 we not know the symbol used, on position 6 we can not distinguished between a '0' or a '1'.

For simplify the graphic presentation, we labelled each input state by its associates sign and to each not input state we labelled it with a number. Lets $$_{\tiny\langle I(a=1), I(a=0), 1, 2, 3, 4, 5, 6\rangle}$$ be the list of states, where $I(a=1)$ and $I(a=0)$ represent the input states, its uncertainty is indexed to the uncertainty on the reading of sign $'a=1'$ and $'a=0'$, and let $4,5,$ and $6$ be the automata output states. Let the state change boolean relations described using the graph presented on figure \ref{transition}, where the arrows labelled with 0 represents the propagation of negative uncertainty and the arrows labelled with 1 represents the propagation of positive uncertainty.
\begin{figure}[h]
\[
\tiny
\xymatrix@=12pt{ I_1(a=1)\ar[r]_1 & 2 \ar[rr]_1&   & 4 \ar[dl]_0 \ar[dr]_1&    \\
              & & 3 \ar[dr]_0 \ar[ul]_1&   & 6 \ar[dl]_0 \\
           I_2(a=0)\ar[r]_1 & 1 \ar[ur]_0&   & 5 &
          }
\]
\caption{Uncertain state transition on an automata.}\label{transition}
\end{figure}
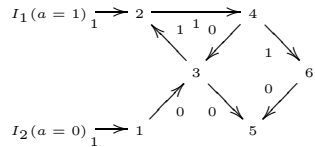
This graph can be codified using two graph adjacency matrices:  $M_0$, for sub-graph having arrows labelled with 0, describing the  state change when a state isn't active, and $M_1$ the adjacency for sub-graphs having arrows labelled with 1, describing behaviour when a state is active.
\[
\tiny
M_0=
\left[
  \begin{array}{cccccccc}
    0 &0& 0 & 0 & 0 & 0 & 0 & 0 \\
    0 &0& 0 & 0 & 0 & 0 & 0 & 0 \\
    0 &0& 0 & 0 & 0 & 0 & 0 & 0 \\
    0 &0& 0 & 0 & 0 & 0 & 0 & 0 \\
    0 &0& 1 & 0 & 0 & 1 & 0 & 0 \\
    0 &0& 0 & 0 & 0 & 0 & 0 & 0 \\
    0 &0& 0 & 0 & 1 & 0 & 0 & 1 \\
    0 &0& 0 & 0 & 0 & 0 & 0 & 0 \\
  \end{array}
\right]
\;\;\;
M_1=
\left[
  \begin{array}{cccccccc}
    0 &0& 0 & 0 & 0 & 0 & 0 & 0 \\
    0 &0& 0 & 0 & 0 & 0 & 0 & 0 \\
    0 &1& 0 & 0 & 0 & 0 & 0 & 0 \\
    1 &0& 0 & 0 & 1 & 0 & 0 & 0 \\
    0 &0& 0 & 0 & 0 & 0 & 0 & 0 \\
    0 &0& 0 & 1 & 0 & 0 & 0 & 0 \\
    0 &0& 0 & 0 & 0 & 0 & 0 & 0 \\
    0 &0& 0 & 0 & 0 & 1 & 0 & 0 \\
  \end{array}
\right]
\]
Executing this automata, for word described on table \ref{word},  and taking the automata initial state uncertain, i.e. making the initial state $_{ e_1=[1/2,1/2,1/2,1/2,1/2,1/2,1/2,1/2]'}$. After reading the first position, we update $e_1$ using the uncertain for each signs on first position,
$_{ e_1=[1,0,1/2,1/2,1/2,1/2,1/2,1/2]'}$,
from it we compute $e_2$ by:
$$
_{e_2= M_0\otimes(\neg e_1)\oplus M_1\otimes e_1=
\left[0, 0, 0, 1, 1, 1/2, 1, 1/2 \right]'.}
$$
Updating the resulting state, since the second mark is 'a=1', we have
$$_{e_2=[1,0,0,1,1,1/2,1,1/2]'}.$$
Using the same procedure
$_{e_3=[1/2,1/2,0,0,1,1,1,1/2,1/2]'}$,
since we not know if in position 3 we have 'a=1' or 'a=0'. Repeating the process for other word positions we have:
\[
\tiny
\begin{array}{ll}
  e_4=[0,1,1/2,1,1,1,1/2,1]' & e_8=[1,0,0,1,3/4,1/2,3/4,3/4]' \\
  e_5=[1/4,3/4,1,1,1/2,1,0,1]' & e_9=[1/2,1/2,0,1,1,1,1/2,1/2]' \\
  e_6=[1/2,1/2,3/4,3/4,0,1,1/2,1]' & e_{10}=[1/4,3/4,1/2,1,1,1,1/2,1]' \\
  e_7=[1,0,1/2,1/2,1/4,3/4,1,1]' & e_{11}=[0,0,2/3,1,1/2,1,0,1]' \\
    & e_{12}=[0,1,1,1/2,1/4,1,1/2,1]'
\end{array}
\]
The automata final state is defined using,
$_{e_{13}=[0,0,1,1/4,0,1/2,3/4,1]'}$, and it is $_{A(w)=[1/2,3/4,1]'}$ since these are the uncertainties in $e_{13}$ for output states $\{4,5,6\}$.
We will interpret $_{A(w)=[1/2,3/4,1]'}$ as the output for automata $A$ when it is executed over the fuzzy string $w$.
\end{exam}

\begin{figure}[h]
\[
\tiny
\begin{array}{ccc}
\xymatrix @=15pt { c_1\ar[rrd]^1 &  c_2\ar[dr]^1 &c_3\ar[d]^1& c_4\ar[dl]_0 & c_5\ar[lld]^0 \\
           &&*+[o][F-]{c_i}&   \\
          }
          &
\xymatrix @=10pt {  &  &   \\
                   &  &   \\
                 \ar[rr] & &   \\
                  &  &   \\
                  &  &
          }
          &
\xymatrix @=15pt { c_1\ar[rrd]^1 & c_2\ar[rd]^1 & c_3\ar[d]^1  & c_4\ar[ld]_{-1} & c_5\ar[lld]^{-1} \\
           \ar@{-}[rr]^2 & &*+[o][F-]{c_i}&   \\
          }
\end{array}
\]
\caption{Translating a neighbor dependence for a state $i$ in a Castro neural network.}\label{actconf}
\end{figure}
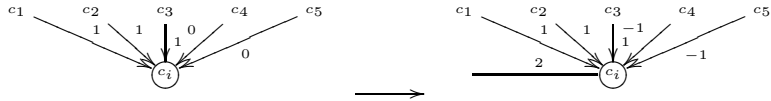

Since in this type of automata transitions are defined using only boolean relations, they can be described using CNNs. Each node depends on a positive or on a negative way from its neighbours, and the relation can be expressed  through a sentence in disjunctive form. We can see this translation from a local configuration to an admissible configuration on figure \ref{actconf}. Note what, neuron bias is defined by the number of arrows labelled with zero. In string-base notation it could be write as
$
_{c_i=c_1\oplus c_2\oplus c_3\oplus \neg c_4\oplus \neg c_5,}
$ which can also be seen as a colimit sentence.

\subsection{Reverse engineering {\L}ukasiewicz automatas}

The extraction of knowledge from data generated by automata execution offers a rich framework to present a simple knowledge integration methodology on {\L}logic. For this we used automata to generate complex artificial data sets.

\begin{figure}[h]
\[
\tiny
\begin{array}{cc}
\xymatrix @=10pt {I(a=1)\ar[r]_1  & A_3 \ar[rr]_0&   & A_5 &    \\
          &   & A_4 \ar[ur]_1 \ar[ul]_0&   & A_7 \ar[ul]_0\\
           I(a=0)\ar[r]_1 & A_2 \ar[ur]_1\ar[rr]_0&   & A_6\ar[ur]_1\ar[ul]_0 &
          }
          &
\xymatrix @=10pt { I(a=1)\ar[r]_1 & A_3 \ar[rr]_0 &   & A_5\ar[dd]_1 &    \\
          &   & A_4 \ar[ur]_1\ar[ul]_0 &   & A_7 \ar[ul]_0\\
           I(a=0)\ar[r]_1 & A_2 \ar[ur]_1\ar[rr]&   & A_6\ar[ur]_1 \ar[ll]_0  \ar[lu]_0&
          }
\end{array}
\]
\caption{Acyclic and cyclic binary automats.}\label{exmp2}
\end{figure}
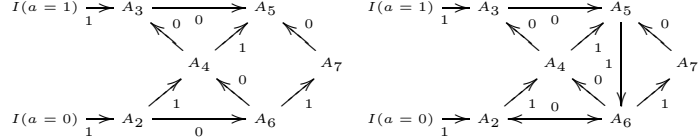

We selected two binary automata described graphically on figure \ref{exmp2}, and constructed data sets using information generated  through automata execution. This was made selecting  possible input configuration with length 6, in a 5-valued {\L}logic. Where we impose a consistence principle in the automata reading sensor: the uncertainty for sign '$s=0$', is the negation of sign '$s=1$' uncertainty. For both automata we generated a data set describing the dependence between states,  relating state uncertainty on iteration $t$ with the state in iteration $t+1$.  With this data set we reverse engineering the automata structure, using Algorithm \ref{RevEng} presented in last section. Trying predict the uncertainty in each state on iteration $t+1$ using the information, about the automata state uncertainty, in iteration $t$ and the input uncertainty. For each prediction task we selected a rule describing the relationships between states. Figure \ref{neuroConf} describes the reverse engineering output for each state, generated using the data produced by the acyclic automata from figure \ref{exmp2}.

\begin{figure}
\begin{center}
\tiny
\begin{tabular}{ccc}
\begin{tabular}{lll}
\tiny
  $\left[
    \begin{array}{cccccccc}
      -1 &  0 & 0 & 0 & 0 & 0 & 0 & 0\\
    \end{array}
  \right]
  $ &$ \left[
       \begin{array}{c}
         1 \\
       \end{array}
     \right]
    $& $\begin{array}{l}
         \neg I(a_t=1) \\
       \end{array}$\\
  $\left[
    \begin{array}{c}
      1 \\
    \end{array}
  \right]
   $&$ \left[
       \begin{array}{c}
         0 \\
       \end{array}
     \right]
    $& $\begin{array}{l}
          i_1 \\
       \end{array}$\\
  $\left[
       \begin{array}{cc}
         1 \\
       \end{array}
     \right] $& $\left[
       \begin{array}{c}
         0 \\
       \end{array}
     \right]$ & $j_1$\\
\end{tabular}
&
\begin{tabular}{lll}
\tiny
  $\left[
    \begin{array}{cccccccc}
      -1 &  0 & 0 & 0 & 1 & 0 & 0 & 0\\
    \end{array}
  \right]
  $ &$ \left[
       \begin{array}{c}
         0 \\
       \end{array}
     \right]
    $& $\begin{array}{l}
         \neg I(a_t=1) \otimes A_4\\
       \end{array}$\\
  $\left[
    \begin{array}{c}
      -1 \\
    \end{array}
  \right]
   $&$ \left[
       \begin{array}{c}
         1 \\
       \end{array}
     \right]
    $& $\begin{array}{l}
         \neg i_1 \\
       \end{array}$\\
  $\left[
       \begin{array}{cc}
         1 \\
       \end{array}
     \right] $& $\left[
       \begin{array}{c}
         0 \\
       \end{array}
     \right]$ & $j_1$\\
\end{tabular}\\
$_{A_2(t+1)= I(a_t=1)}$&$_{A_3(t+1)=\neg(\neg I(a_t=1)\oplus A_4(t))}$\\
\\
\begin{tabular}{lll}
\tiny
  $\left[
    \begin{array}{cccccccc}
      0 &   0 & -1 & 0 & 0 & 0 & 1 & 0\\
    \end{array}
  \right]
  $ &$ \left[
       \begin{array}{c}
         0 \\
       \end{array}
     \right]
    $& $\begin{array}{l}
         \neg A_2 \oplus A_6\\
       \end{array}$\\
  $\left[
    \begin{array}{c}
      -1\\
    \end{array}
  \right]
   $&$ \left[
       \begin{array}{c}
         1 \\
       \end{array}
     \right]
    $& $\begin{array}{l}
         \neg i_1 \\
       \end{array}$\\
  $\left[
       \begin{array}{cc}
         1 \\
       \end{array}
     \right] $& $\left[
       \begin{array}{c}
         0 \\
       \end{array}
     \right]$ & $j_1$\\
\end{tabular}
&
\begin{tabular}{lll}
\tiny
  $\left[
    \begin{array}{cccccccc}
      0 &  0 & 0 & -1 & 1 & 0 & 1 & -1\\
    \end{array}
  \right]
  $ &$ \left[
       \begin{array}{c}
         2 \\
       \end{array}
     \right]
    $& $\begin{array}{l}
          \neg A_3\oplus A_4\oplus A_6 \oplus \neg A_7 \\
       \end{array}$\\
  $\left[
    \begin{array}{c}
      1 \\
    \end{array}
  \right]
   $&$ \left[
       \begin{array}{c}
         0 \\
       \end{array}
     \right]
    $& $\begin{array}{l}
         i_1 \\
       \end{array}$\\
  $\left[
       \begin{array}{cc}
         1 \\
       \end{array}
     \right] $& $\left[
       \begin{array}{c}
         0 \\
       \end{array}
     \right]$ & $j_1$\\
\end{tabular}\\
$_{A_4(t+1)=\neg (\neg A_2(t) \oplus A_6(t))}$&$_{A_5(t+1)\sim_{0.97}\neg A_3(t)\oplus A_4(t)\oplus A_6(t) \oplus \neg A7}$\\
\\
\begin{tabular}{lll}
\tiny
  $\left[
    \begin{array}{cccccccc}
      0 & 0 & -1 & 0 & 0 & 0 & 0 & 0\\
    \end{array}
  \right]
  $ &$ \left[
       \begin{array}{c}
         1 \\
       \end{array}
     \right]
    $& $\begin{array}{l}
         \neg A_2 \\
       \end{array}$\\
  $\left[
    \begin{array}{c}
      1 \\
    \end{array}
  \right]
   $&$ \left[
       \begin{array}{c}
         0 \\
       \end{array}
     \right]
    $& $\begin{array}{l}
         i_1 \\
       \end{array}$\\
  $\left[
       \begin{array}{cc}
         1 \\
       \end{array}
     \right] $& $\left[
       \begin{array}{c}
         0 \\
       \end{array}
     \right]$ & $j_1$\\
\end{tabular}
&
\begin{tabular}{lll}
  $\left[
    \begin{array}{cccccccc}
      0 &  0 & 0 & 0 & 0 & 0& -1 & 0 \\
    \end{array}
  \right]
  $ &$ \left[
       \begin{array}{c}
         1 \\
       \end{array}
     \right]
    $& $\begin{array}{l}
         \neg A_6 \\
       \end{array}$\\
  $\left[
    \begin{array}{c}
      -1 \\
    \end{array}
  \right]
   $&$ \left[
       \begin{array}{c}
         1 \\
       \end{array}
     \right]
    $& $\begin{array}{l}
         \neg i_1 \\
       \end{array}$\\
  $\left[
       \begin{array}{cc}
         1 \\
       \end{array}
     \right] $& $\left[
       \begin{array}{c}
         0 \\
       \end{array}
     \right]$ & $j_1$\\
\end{tabular}
\\
$_{A_6(t+1)=\neg A_2(t)}$&$_{A_7(t+1)=A_6(t)}$
\\
\end{tabular}
\end{center}
\caption{Neuro network configuration extract for each non input state.}\label{neuroConf}
\end{figure}

Each configuration can be expressed or approximate using a string-based presentation. Each of this formulas is interpreted as knowledge extracted from different views of the data set. The set of all extracted formulas (equalities) we call a \emph{theory}, and can be codified as an specification system. For acyclic automata, on figure \ref{exmp2}, and translating configurations presented in figure \ref{neuroConf}, we have the theory:
 \[
 _{T_0=\{A_2(t+1)=I(a_t=1),\;A_3(t+1)=\neg(\neg I(a_t=1)\oplus A_4(t)),\;A_4(t+1)=\neg (\neg A_2(t) \oplus A_6(t)), }
 \]
 \[_{A_5(t+1)\sim_{0.97}\neg A_3(t)\oplus A_4(t)\oplus A_6(t) \oplus \neg A_7(t),\; A_6(t+1)=\neg A_2(t),\;A_7(t+1)=A_6(t)\}}
 \]
This symbolic description allows forecast the automata behaviour. In this case theory $T_0$ isn't a prefect automata description, since we only have a approximation to automata beaver on state $A_5$. The consistence level of a equational theory $T$, with a model $D$, is describes by the disjunction of each similarity level associated to each equality in $T$. In the case of equational theory $T_0$, we have a consistence level of $0.97$. Given a generic word $w=s_0s_1s_2s_3s_4$ in the interaction $t=5$, the uncertainty on state $A_4$ depends on signs from positions $s_2$ and $s_3$.
\begin{center}
\tiny
$
  \begin{array}{rcl}
  A_4(5) & = & A_2(4)\otimes \neg A_6(4) \\
         & = & I(s_3=1)\otimes \neg A_2(3) \\
         & = & I(s_3=1)\otimes I(s_2=1)  \\
\end{array}
$
\end{center}
In interaction $t=6$, the uncertainty on state $A_5$, is given by:
\begin{center}
\tiny
$
  \begin{array}{rcl}
  A_5(6) & \sim_{0.97} & \neg A_3(5)\otimes A_4(5)\otimes A_6(5) \otimes \neg A_7(5)\\
         & \sim & \neg I(s_2=1)\oplus ( I(s_3=1)\otimes I(s_2=1))\oplus \neg A_3(4) \oplus \neg A_6(4) \\
         & \sim & \neg I(s_2=1)\oplus I(s_3=1)\oplus \neg I(s_3=1) \oplus A_4(3) \oplus A_3(3)\\
         & \sim & 1\\
\end{array}
$
\end{center}

From a functional point of view, knowledge integrated in theory $T_0$, are interpreted using the first {\L}ukasiewicz neural network on figure \ref{knowledgeInteg}. This NN configuration results from integrating NNs presented on figure \ref{neuroConf}. The resulting NN have 6 outputs, one for each non input state.
\begin{figure}
\[
\tiny
\begin{tabular}{ccc}
\begin{tabular}{lll}
  $\left[
    \begin{array}{ccccccc}
      -1 &  0 & 0 & 0 & 0 & 0 & 0 \\
      -1 &  0 & 0 & 1 & 0 & 0 & 0 \\
      0 &  0 & -1 & 0 & 0 & 1 & 0 \\
      0 &  0 & -1 & 1 & 0 & 1 & -1 \\
      0 &  0 & -1 & 0 & 0 & 0 & 0 \\
      0 &  0 & 0 & 0 & 0 & -1 & 0 \\
    \end{array}
  \right]
  $ &$ \left[
       \begin{array}{c}
         1 \\
         0 \\
         0 \\
         2 \\
         1 \\
         1 \\
       \end{array}
     \right]
    $& $\begin{array}{l}
         I(a=1) \\
         \neg I(a=1) \\
         \neg A_2 \oplus A_6 \\
         \neg A_3 \oplus A_4\oplus A_6\oplus \neg A_7 \\
         \neg A_2 \\
         \neg A_7 \\
       \end{array}$\\
  $\left[
    \begin{array}{ccccccc}
      1 & 0& 0& 0& 0& 0 \\
      0& -1& 0& 0& 0& 0 \\
      0& 0& -1& 0& 0& 0 \\
      0& 0& 0&  1& 0& 0 \\
      0& 0& 0&  0& 1& 0 \\
      0& 0& 0&  0& 0& -1 \\
    \end{array}
  \right]
   $&$ \left[
       \begin{array}{c}
         1 \\
         0 \\
         1 \\
         0 \\
         0 \\
         0 \\
       \end{array}
     \right]
    $& $\begin{array}{l}
         i_1 \\
         \neg i_2 \\
         \neg i_3 \\
         i_4 \\
         i_5 \\
         \neg i_6 \\
       \end{array}$\\
\end{tabular}
&
\;\;\;
&
\begin{tabular}{lll}
  $\left[
    \begin{array}{cccccccc}
      0 &  1 & 0 & 0 & 0 & 0 & -1 & 0\\
      1 &  0 & 0 & 0 & -1 & 0 & 0 & 0\\
      -1 &  0 & 0 & 0 & 1 & 0 & 0 & 0\\
      0 &  0 & -1 & 0 & 0 & 0 & 1 & 0\\
      0 &  0 & 0 & -1 & 1 & 0 & 0 & -1\\
      0 &  0 & 0 & 0 & 0 & 0 & 1 & 0\\
    \end{array}
  \right]
  $ &$ \left[
       \begin{array}{c}
         1 \\
         1 \\
         0 \\
         0 \\
         2 \\
         0 \\
       \end{array}
     \right]
    $& $\begin{array}{l}
         I(a=0)\oplus \neg A_6 \\
         I(a=1)\oplus \neg A_4 \\
         \neg I(a=1) \oplus A_4 \\
         \neg A_2\oplus A_6  \\
         \neg A_3 \oplus A_4 \oplus \neg A_7\\
         A_6 \\
       \end{array}$\\
  $\left[
    \begin{array}{ccccccc}
      1 & 0& 0& 0& 0& 0 \\
      0& 1& 0& 0& 0& 0 \\
      0& 0& 1& 0& 0& 0 \\
      0& 0& 0& -1& 0& 0 \\
      0& 0& 0& 0& 1& 0 \\
      0& 0& 0& 0& 0& 1 \\
    \end{array}
  \right]
   $&$ \left[
       \begin{array}{c}
         0 \\
         0 \\
         1 \\
         1 \\
         0 \\
         0 \\
       \end{array}
     \right]
    $& $\begin{array}{l}
         i_1 \\
         i_2 \\
         \neg i_3 \\
         i_4 \\
         i_5 \\
       \end{array}$
\end{tabular}
\end{tabular}
\]
\caption{Integration of knowledge extracted from automats presented on figure \ref{exmp2} using NNs.}\label{knowledgeInteg}
\end{figure}

Integrating configuration representing knowledge in an NN usually introduce redundancies. In the sense that the same or similar information is codified by different configurations in different locations. This can be minimized by regularizing the produced NN using "Optimal Brain Surgeon" method \cite{Hassibi93}.

In the other direction, every formula in {\L}ukasiewicz logic can be used to describe a state uncertainty. By this we mean that, for every formula we can define an automata with a state having by uncertainty level the result of formula evaluation, after some interactions. For this the automata must have by input a string, having by first position the uncertainty on formula arguments, followed by marks with uncertainty zero for every possible sign. The number of positions in input sting must be equal to the formula parsing tree height.

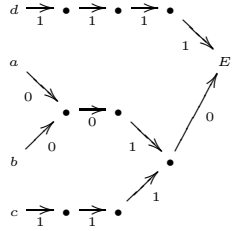
\begin{figure}
\[
\tiny
\xymatrix @=10pt{  d\ar[r]_1  & \bullet\ar[r]_1  &\bullet\ar[r]_1   &\bullet\ar[rd]_1  &  & \\
                   a\ar[rd]_0 &         &         &         &E  & \\
                            & \bullet\ar[r]_0  &\bullet\ar[rd]_1  &         &  & \\
                   b\ar[ru]_0 &         &         &\bullet\ar[ruu]_0  &  & \\
                   c\ar[r]_1  & \bullet\ar[r]_1  &\bullet\ar[ru]_1 &         &  & \\
          }
\]
\caption{Interpretation using an automata for a first-order formula.}\label{forAut}
\end{figure}

Injecting a formula in an automata structure is a simple process. For this we must note what every formula can be expressed in the disjunctive normal form, i.e. using only the connectives disjunction and negation. For instance the formula $((a\otimes b) \oplus c)\rightarrow d$, is equivalent to $\neg(\neg(\neg a\oplus \neg b) \oplus c)\oplus d$, evaluated as the uncertainty on state $E$, after 4 interaction, using automata describe in figure \ref{forAut}. Note that, this defines a concept view $_{R:O\times \{a,b,c,d\}\rightharpoonup \{E\}}$.

\section{\uppercase{Knowledge integration via specification systems}}

The integration of knowledge using connective models is very restrictive, and difficult to be used directory by domain experts. However, it can be useful to deploy analytic models, or on procedures for redundancy minimization in a knowledge base.

We want to present specification systems as the suitable framework for symbolic knowledge integration. Given a semiotic $(S,M)$ our goal is to complete the semiotic by enriching specification $S$, such that the new specification $S'$ is a better description for $M$. This is made by enriching structure $S$ with knowledge
extracted from different views on $M$.

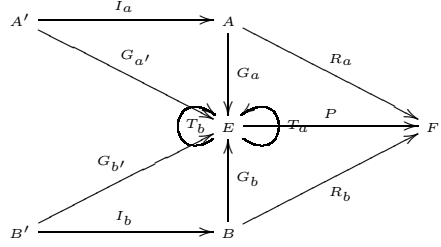
\begin{figure}
\[
\tiny
\xymatrix @=30pt{A' \ar[rr]^{I_a}\ar[rrd]^{G_{a'}}& &A \ar[rrd]^{R_a} \ar[d]^{G_a} & &         \\
          & &E\ar@(dr,ur)[]_{T_a}\ar@(dl,ul)[]_{T_b}  \ar[rr]^{P}& &   F      \\
          B' \ar[rr]^{I_b}\ar[rru]^{G_{b'}}& &B \ar[rru]_{R_b} \ar[u]_{G_b} & &         \\
          }
\]
\caption{Graphical presentation.}\label{GraphP}
\end{figure}

For that, in this section, we extract knowledge from a structure construct using fuzzy relations. This relations are known to be defined using fuzzy automata, but this fact is hidden to the completion methodology.

Figure \ref{GraphP} describe the UoD structure, where concept views $G_a$ and $G_b$ are known to be described, respectively, by the acyclic and cyclic automata on  figure \ref{exmp2}, where $A$ and $B$ are sets of input strings and $E$ is the set of automata possible states. This relations are described by two data sets with 15625 cases, using 14 attributes.  Relations $T_a$ and $T_b$ are defined by the state transformation relations, codified on two data sets with 15625 cases, using 16 attributes, each. In the specification, views $R_a$ and $B_b$ denote the set of automata output states, given through the selection by $F$ of automata final states in $E$, satisfying: $R_a=G_a\otimes F$ and $R_a=G_b\otimes F$.

Generically, we can formalize this structure using the following string-based relational specification:
\begin{center}
\tiny
\begin{tabular}{cc}
\begin{tabular}{rcl}
   \%Nodes:\\
   $I$&:&$\{0,1\};$ \\
   $A,B,A',B'$&:&I,I,I,I,I,I,I,I;\\
   $F$&:&I;\\
   $E$&:&$\{A_0,A_1,A_2,A_3,A_4,A_5,A_6,A_7\};$\\
   \\
   \%Arrows:\\
   $R_a$&:&$\{A\rightharpoonup F\};$ \\
   $G_a$&:&$\{A\rightharpoonup E\};$ \\
   $R_b$&:&$\{B\rightharpoonup F\}; $\\
   $G_b$&:&$\{B\rightharpoonup F\}; $\\
   $P$&:&$\{E\rightharpoonup F\}; $\\
   $T_a,T_b$&:&$\{E\rightharpoonup E\}; $\\
   $I_a$&:&$\{A'\rightharpoonup A;$\\
          &  &\begin{tabular}{rll}
             $\Gamma_a:\{$ &$A'$;         & \\
                & $\Gamma_a$ : similarity;\\
          & \} \\
            \end{tabular}\\
         && $I_a$ : is\_a$(\Gamma_a)$; \\
            && $\};$ \\
   $I_b$&:&$\{B'\rightharpoonup B;$\\
           & &\begin{tabular}{rll}
             $\Gamma_b:\{$ &$B'$;         & \\
                & $\Gamma_b$ : similarity;\\
          & \} \\
            \end{tabular}\\
        & &$I_b$ : is\_a$(\Gamma_b)$; \\
            && $\};$ \\
   \\
   \end{tabular}
   &
   \begin{tabular}{rcl}
   \%Commutative diagrams:\\
   $D_1$&:&$\{A\rightharpoonup F;$ \\
        &  &$D_1:R_a\otimes G_a\otimes P$; \\
        & &\}\\
   $[D_1]$\\
   \\
   $D_2$&:&$\{B\rightharpoonup F;$ \\
        &  &$D_2:R_b\otimes G_a\otimes P$; \\
        & &\}\\
   $[D_2]$\\
   \\
   $D_3$&:&$\{A'\rightharpoonup E;$ \\
        &  &$D_3:I_a\otimes G_a\otimes G_{a'}$; \\
        & &\}\\
   $[D_3]$\\
   \\
   $D_4$&:&$\{B'\rightharpoonup E;$ \\
        &  &$D_4:I_b\otimes G_b\otimes G_{b'}$; \\
        & &\}\\
   $[D_4]$\\
   \\
   \end{tabular}   \\
   \end{tabular}
\end{center}

This specification have the model, described by a set of data sets generated using fuzzy automata. This data sets are interpretations for each of the arrows used on the specification. Using the extraction process, described in the last section, we can extract rules with insights about the UoD structure.

The knowledge extracted from crystallized rules from CNN, when trained over views $G_a$ and $G_b$, describe the relation between sign uncertainty on input strings, with length 6, and the uncertainty for each state after the string have been read. Since extraction is made in a supervising learning context, the process is oriented for the perdition of each automata state, using  input string. Every obtain configuration, in this cases was representable, simplifying the translation for the string-based rules presented bellow for the knowledge base enrichment.

\begin{center}
\tiny
\begin{tabular}{rcl}
   $G_a$&:&$\{A\rightharpoonup E;$\\
   & &$\begin{array}{ccl}
     G_a(s_1,s_2,s_3,s_4,s_5,s6) & : & \\
      & : & A_0 =  s_6; \\
      & : & A_1 = \neg s_6; \\
      & : & A_2 = \neg s_5; \\
      & : & A_3 =  \neg(\neg s_6)\oplus (s_3 \otimes s_4);\\
      & : & A_4 =  \neg(s_4\oplus s_5);\\
      & : & A_5 = \neg (s_3 \oplus s_4) \oplus  (\neg s_2\otimes s_3\otimes \neg s_5);\\
      & : & A_6 = s5;\\
      & : & A_7 = s4;\\
   \end{array}$\\
   & &$\};$ \\
   $G_b$&:&$\{B\rightharpoonup E; $\\
   & &$\begin{array}{ccl}
     G_b(s_1,s_2,s_3,s_4,s_5,s_6) & : &  \\
      & : & A_0=  s_6; \\
      & : & A_1= \neg s_6; \\
      & : & A_2\sim_{0.9391}  (s_2\otimes \neg s_4)\oplus \neg s_6; \\
      & : & A_3\sim_{0.9961}  \neg(\neg s_4 \otimes \neg s_6);\\
      & : & A_4\sim_{0.9785}  \neg (s_2\otimes s_3 \otimes \neg s_4);\\
      & : & A_5\sim_{0.9947}  \neg s_4 \oplus ( \neg s_3\otimes \neg s_5);\\
      & : & A_6\sim_{0.9429}  \neg (s_3\otimes \neg s_5);\\
      & : & A_7\sim_{0.9767}  \neg s_2\oplus \neg s_3 \oplus s_4;\\
   \end{array}$\\
   & &$\};$ \\
\end{tabular}
\end{center}

The knowledge presented on figure \ref{knowledgeInteg} describes the relation between automata states. It was generated predicting the uncertainty in each state in interaction $i+1$ using automata state uncertainty in interaction $i$ and input signs uncertainty. We present bellow the best rules generated by the extraction process for each task.
\begin{center}
\tiny
\begin{tabular}{rcl}
   $T_a$&:&$\{E\rightharpoonup E$;\\
   & & $\begin{array}{ccl}
     T_a(A_0,A_1,A_2,A_3,A_4,A_5,A_6,A_7) & : &  \\
      & : & A_2 = \neg A_0; \\
      & : & A_3 = \neg(\neg A_0 \oplus A_4); \\
      & : & A_4 = \neg(\neg A_2 \oplus A_6); \\
      & : & A_5\sim_{0.9747}  \neg A_3\oplus A_4\oplus A_6 \oplus \neg A_7);\\
      & : & A_6 =  \neg A_3;\\
      & : & A_7 =  A_6;\\
   \end{array}$\\
      & &\}\\
   $T_b$&:&$\{E\rightharpoonup E$;\\
   & & $\begin{array}{ccl}
     T_b(A_0,A_1,A_2,A_3,A_4,A_5,A_6,A_7) & : &  \\
      & : & A_2 =  A_1 \oplus \neg A_6; \\
      & : & A_3 = \neg A_0 \otimes A_4; \\
      & : & A_4 = ;\\
      & : & A_5 = \neg(\neg A_2 \otimes A_6); \\
      & : & A_6\sim_{0.9993}  \neg A_3\oplus A_4 \oplus \neg A_7);\\
      & : & A_7 =  A_6;\\
   \end{array}$\\
      & &\}\\

\end{tabular}
\end{center}
We can improve the description of our UoD, presenting constrains valid on queries to the specification model. For example, from query defined using equalizer $G_{a'\otimes b'}:A'\times B'\rightharpoonup E$ for view $G_{a'}$ and $G_{b'}$, we can generate new insights in the model structure, which can be used in knowledge base enrichment.
\begin{center}
\tiny
\begin{tabular}{rcl}
   \%Limit sentences:\\
   $D_3$&:&$\{A',B'\rightharpoonup E;$ \\
        &  &$D_3:=G_{a'}\otimes G_{b'}$; \\
        & &\}\\
   $G_{a\otimes b}$&:&$\{ A,B,E;$ \\
      & & $G_{a\otimes b}:Lim\;D_3$;\\
      & & $\begin{array}{ccl}
     G_{a\otimes b}(s_1,\ldots,s_6,s'_1,\ldots,s'_6) & : &  \\
      & : & A_0= s_6 \otimes s'_6; \\
      & : & A_1= \neg(s_6 \oplus s'_6); \\
      & : & A_2\sim_{0.9878} (s_6 \oplus \neg s'_6)\oplus (\neg s_6 \otimes s'_2\otimes s'_3\otimes \neg s'_4\otimes s'_6); \\
      & : & A_3\sim_{0.9873}\neg(\neg s_6\oplus \neg s'_4)\oplus(s_6\otimes s'_6);\\
      & : & A_4\sim_{0.9869}(\neg s_5\oplus \neg s'_2\oplus s_4)\oplus(s'_1\oplus s'_3\oplus \neg s'_5)\oplus(s_5\oplus s'_5);\\
      & : & A_5\sim_{0.9609}((\neg s'_4\oplus \neg s'_5)\oplus(s'_3\otimes s'_4)\oplus\neg(s_4\oplus\neg s_5))\otimes\\
      &  & \otimes\neg((s'_3\otimes s'_4)\otimes\neg(s_4\oplus s_5));\\
      & : & A_6=(s_4\oplus s_5)\otimes\neg(s'_3\otimes s'_4\otimes \neg s'_5)\otimes(s'_2\oplus s'_3\oplus s'_5);\\
      & : & A_7\sim_{0.9526}\neg s'_2\oplus \neg s'_3\oplus s_4;\\
   \end{array}$\\
      & &\}
\end{tabular}
\end{center}

\section{\uppercase{Conclusions}}
This methodology to codify and extract symbolic knowledge from a NN is very simple and efficient for the extraction of comprehensible rules from medium-sized data sets. It is, moreover, very sensible to attribute relevance.

In the theoretical point of view it is particularly interesting that restricting the values assumed by neurons weights restrict the information propagation in the network, thus allowing the emergence of patterns in the neuronal network structure. For the case of linear neuronal networks, having by activation function the identity truncate to 0 and 1, these structures are characterized by the occurrence of patterns in neuron configuration  directly presentable as formulas in {\L}logic.

The application of procedures like the one above, on information systems, generates grates amount of information. We organize this information in a specification systems using a relational language. And we propose this language as an Interface Layer for AI. Here, classic graphical models like Bayesian and Markov networks have to some extent played the part of an interface layer, but one with a limited range having insufficiently expressive for general AI \cite{Domingos06}.

\bibliographystyle{splncs}
{\tiny
\bibliography{multibib}}
\end{document}